\pdfoutput=1

\documentclass[11pt]{article}

\usepackage{emnlp2022}

\usepackage{times}
\usepackage{latexsym}
\usepackage{tabularx}
\usepackage{xcolor}
\usepackage{soul}
\usepackage{amsmath}
\usepackage{amssymb}
\usepackage{booktabs}
\usepackage{tikz}
\usepackage{pgfplots}
\usepackage{caption}
\usepackage{subcaption}

\pgfplotsset{compat=1.17} 
\usepackage[T1]{fontenc}

\usepackage[utf8]{inputenc}

\usepackage{microtype}

\newcommand{\push}{\textsc{Push}}
\newcommand{\adv}{\textsc{Advance}}
\newcommand{\pop}{\textsc{Pop}}
\newcommand{\peek}{\textsc{Peek}}
\newcommand{\coref}{\textsc{Coref}}

\newcommand{\muc}{$\textit{MUC}$}
\newcommand{\bcub}{$\textit{B}^3$}
\newcommand{\ceaf}{$\textit{CEAF}_{\phi_4}$}
\newcommand{\mtop}{\mathbf{m}_{i^*}}
\DeclareMathOperator{\mlp}{MLP}
\DeclareMathOperator{\lstm}{LSTM}
\DeclareMathOperator{\stacklstm}{StackLSTM}
\DeclareMathOperator*{\argmax}{arg\,max}

\usetikzlibrary{matrix}
\tikzstyle{component} = [rectangle, thick, rounded corners, minimum width=0.7cm, minimum height=2cm,draw=black, fill=bluex!20]
\tikzstyle{component2} = [rectangle, thick, rounded corners, minimum width=1.2cm, minimum height=2cm,draw=black, fill=purple!40]
\tikzstyle{arrow} = [thick,->,>=stealth]
\definecolor{redx}{RGB}{228,26,28}
\definecolor{bluex}{RGB}{55,126,184}
\definecolor{greenx}{RGB}{77,175,74}
\definecolor{purplex}{RGB}{152,78,163}
\definecolor{orangex}{RGB}{255,127,0}

\newcommand{\emldisplay}[2]{\texttt{\href{mailto:#1}{#2}}}
\newcommand{\eml}[1]{\emldisplay{#1}{#1}}

\title{Sentence-Incremental Neural Coreference Resolution}

\author{Matt Grenander \quad Shay B. Cohen \quad Mark Steedman \\
        School of Informatics \\
        University of Edinburgh \\
        \eml{matt.grenander@ed.ac.uk} \\
        \eml{scohen@inf.ed.ac.uk}, \eml{m.steedman@ed.ac.uk} \\        
}

\begin{document}
\long\def\/*#1*/{}
\maketitle

\begin{abstract}
We propose a sentence-incremental neural coreference resolution system which incrementally builds clusters after marking mention boundaries in a shift-reduce method.
The system is aimed at bridging two recent approaches at coreference resolution:
(1) state-of-the-art non-incremental models that incur quadratic complexity in document length with high computational cost, and
(2) memory network-based models which operate incrementally but do not generalize beyond pronouns.
For comparison, we simulate an incremental setting by constraining non-incremental systems to form partial coreference chains before observing new sentences.
In this setting, our system outperforms comparable state-of-the-art methods by 2 F1 on OntoNotes and 7 F1 on the CODI-CRAC 2021 corpus.
In a conventional coreference setup, our system achieves 76.3 F1 on OntoNotes and 45.8 F1 on CODI-CRAC 2021, which is comparable to state-of-the-art baselines.
We also analyze variations of our system and show that the degree of incrementality in the encoder has a surprisingly large effect on the resulting performance.\footnote{\raggedright Code is available at: \href{https://github.com/mgrenander/sentence-incremental-coref}{https://github.com/mgrenander/sentence-incremental-coref}}
\end{abstract}

\section{Introduction}

\textit{Coreference Resolution} (CR) is a task in which a system detects and resolves linguistic expressions that refer to the same entity. 
It is typically performed in two steps: in \textit{mention detection}, the model predicts which expressions are referential, and in \textit{mention clustering}, the model computes each mention's antecedent.
Many recently proposed systems follow a mention-pair formulation from \citet{lee-etal-2017-end}, in which all possible spans are ranked and then scored against each other.
In particular, methods that augment this approach with large, pre-trained language models achieve state-of-the-art results \cite{joshi-etal-2019-bert, joshi-etal-2020-spanbert}.

Despite impressive performance, these methods are computationally demanding.
For a text with $n$ tokens, they will score up to $O(n^2)$ spans, followed by up to $O(n^4)$ span comparisons.
They also process documents non-incrementally, requiring access to the entire document before processing can begin.
These properties present challenges when insufficient computational resources are available, or when the task setup is incremental, such as in dialogue (e.g. \citealt{khosla-etal-2021-codi}).
From a cognitive perspective, these methods are also unappealing because research on ``garden-path'' effects show that humans resolve referring expressions incrementally \cite{altmann}.

\begin{figure}[t]
    \centering
    \begin{tabularx}{\linewidth}{|X|}
    \hline
        In 2004, on the Waterfront Promenade originally constructed for viewing only the scenery of  \colorbox{yellow}{Hong Kong Island} and Victoria Harbor, the \colorbox{green}{Hong Kong Tourism Board} also constructed the Avenue of Stars, memorializing \colorbox{cyan}{Hong Kong's 100-year film history}.
    \\\hline
    \end{tabularx}
    \caption{An example from the OntoNotes dataset which highlights the need for incremental systems to identify spans rather than tokens as mentions. The mentions cannot be resolved solely from the prefix `Hong Kong', and the clustering decision should be delayed until the full mention is observed.}
    \label{fig:problem_example}
\end{figure}

These drawbacks have led to renewed interest in incremental coreference resolution systems, in which document tokens are processed sequentially. 
Some recent approaches use memory networks to track entities in differentiable memory cells \cite{liu-etal-2019-referential, toshniwal-etal-2020-petra}. 
These models demonstrate proficiency at proper name and pronoun resolution \cite{webster-etal-2018-mind}.
However, they seem unlikely to generalize to more complicated coreference tasks due to a strict interpretation of incrementality. 
Both \citet{liu-etal-2019-referential} and \citet{toshniwal-etal-2020-petra} resolve mentions word-by-word, making coreference decisions possibly before the full noun phrase has been observed.
The approach is adequate for proper names and pronouns, but it may fail to distinguish entities who share the same phrase prefix.
For example, in Figure \ref{fig:problem_example}, three mentions all begin with `Hong Kong', though all belong to separate entities.
In this case, it is difficult to see how a system using word-level predictions would resolve these mentions to different entities.


Motivated by this recent work, we propose a new system that processes a document incrementally at the sentence-level, creating and updating coreference clusters after each sentence is observed.
The system addresses deficiencies in memory network-based approaches by delaying mention clustering decisions until the full mention has been observed.
These goals are achieved through a novel mention detector based on shift-reduce parsing, which identifies mentions by marking left and right mention boundaries.
Identified mention candidates are then passed to an online mention clustering model similar to \citet{toshniwal-etal-2020-learning} and \citet{xia-etal-2020-incremental}.
The model proposes a linear number of spans per sentence, reducing computational requirements and maintaining more cognitive plausibility compared to non-incremental methods.

In order to compare non-incremental and incremental systems on equal footing, we propose a new \textbf{sentence-incremental} evaluation setting.
In this setting, systems receive sentences incrementally and must form partial coreference chains before observing the next sentence.
This setting mimics human coreference processing more closely, and is a more suitable evaluation setting for downstream tasks in which full document access is generally not available (e.g. for dialogue \cite{andreas-etal-2020-task}).

Using the sentence-incremental setting, we demonstrate that our model outperforms comparable systems adapted from partly incremental methods \cite{xia-etal-2020-incremental} across two corpora, the OntoNotes dataset \cite{pradhan-etal-2012-conll} and the recently released CODI-CRAC 2021 corpus \cite{khosla-etal-2021-codi}. 
Moreover, we show that in a conventional evaluation setting, where the model can access the entire document, our system retains close to state-of-the-art performance.
However, the sentence-incremental setting is substantially outperformed by non-sentence-incremental systems.
Analyzing the difference between these two settings reveals that the encoder is heavily dependent on how many sentences it can observe at a time.
The analysis suggests better representations of the entities and their context may improve performance in the sentence-incremental setting.
Nevertheless, our results provides new state-of-the-art baselines for sentence-incremental evaluation.




\section{Related Work}
Non-incremental mention-pair models have dominated the field in recent years, with many following the formulation presented by \citet{lee-etal-2017-end}.
Several extensions have led to performance improvements, such as adding higher-order inference \cite{lee-etal-2018-higher}, and replacing the encoder with BERT and SpanBERT \cite{joshi-etal-2019-bert, joshi-etal-2020-spanbert}.
Extensions to this approach have looked at reformulating the problem as question-answering \cite{wu-etal-2020-corefqa}, simplifying span representations  \cite{kirstain-etal-2021-coreference}, and incorporating coherence signals from centering theory \cite{chai-strube-2022-incorporating}.
Although our work is orthogonal to this line of research, we compare our system against this type of non-incremental model.

\citet{toshniwal-etal-2020-learning} and \citet{xia-etal-2020-incremental} adapt the non-incremental system of \newcite{joshi-etal-2020-spanbert} so that mention clustering is performed incrementally.
Their resulting models achieve similar performance to the original non-incremental one. 
However, in their formulation, document encoding, mention detection and certain clustering decisions still fully depend on \newcite{joshi-etal-2020-spanbert}.
The resulting model still requires access to the full document in order to compute coreference chains. 
\citet{yu-etal-2020-cluster} similarly present an incremental mention clustering approach where mention detection is performed non-incrementally as \citet{lee-etal-2017-end}.

Memory network-based approaches identify co-referring expressions by writing and updating entities into cells within a fixed-length memory \cite{liu-etal-2019-referential, toshniwal-etal-2020-petra}.
These models demonstrate how fully incremental coreference systems can be achieved. 
However, the formulation operates on token-level predictions, and does not easily extend to either nested mentions or certain multi-token mentions (e.g. in Figure \ref{fig:problem_example}).

Cross-document coreference resolution (CDCR) requires systems to compute coreference chains across documents, raising scalability challenges as the number of documents increases. 
Given these challenges, incremental CDCR systems are crucial \cite{allaway-etal-2021-sequential, logan-iv-etal-2021-benchmarking} due to lower memory requirements.
However, these works are not directly comparable to ours since they assume gold mentions are provided as input.

Other, earlier, incremental coreference systems also often ignore or diminish the role of mention detection.
For example, \citet{webster-curran-2014-limited} use an external parser for mention detection, requiring an additional model.
\citet{klenner-tuggener-2011-incremental} assume gold mentions as input.

Recently, \citet{liu-etal-2022-transition} propose a coreference resolution system based a seq2seq formulation using hidden variables.
Although their focus is on adding structure to seq2seq models, their system can also be viewed as transition-based like ours.

Our incremental mention detector bears similarities to certain models for nested named-entity recognition (NER). 
In particular, \citet{wang-etal-2018-neural-transition} present an incremental neural model for nested NER based on a shift-reduce algorithm.
Their deduction rules differ greatly from ours as they model mention spans using complete binary trees, and are aimed at NER rather than mention detection.

Recent work has also explored incremental transformer architectures
\cite{katharopoulos_et_al_2020, kasai-etal-2021-finetuning},
and adapting these architectures to NLU tasks (though not coreference resolution) \cite{madureira-schlangen-2020-incremental, kahardipraja-etal-2021-towards}.
In this work, we focus on the simpler sentence-incremental setting, believing it to be sufficient for downstream tasks.

\section{Method}
Given a document, the goal is to output a set of clusters $\mathcal{C}=\{\mathcal{C}_1,\dots,\mathcal{C}_K\}$, where mentions within each cluster are co-referring. 
We assume mentions may be nested but otherwise do not overlap. 
This assumption allows us to model mentions using a method analogous to shift-reduce, where shifting corresponds to either incrementing the buffer index or marking a left mention boundary, and reducing corresponds to marking a right boundary and resolving the mention to an entity cluster.

\subsection{Shift-Reduce Framework}

The main idea is to mark mention boundaries using \push, \pop\ or \peek\ actions, or to pass over a non-boundary token with the \adv\ action. 
After \pop\ or \peek\ actions, a mention candidate is created using the current top-of-stack and buffer elements.
The resulting mention candidate is then either resolved to an existing cluster or initialized as a new entity cluster.

We represent the state as $[S, i, A, \mathcal{C}]$, where $S$ is the stack, $i$ is the buffer index, $A$ is the action history and $\mathcal{C}$ is the current set of clusters. At each time step, one of four actions is taken:
\begin{itemize}
    \item \push: Place the word at buffer index $i$ on top of the stack, marking a left mention boundary.
    \item \adv: Move the buffer index forward.
    \item \pop: Remove the top element from $S$ and create a mention candidate using this element and the current buffer element. Score the candidate against existing clusters and resolve it (or create a new cluster). 
    \item \peek: Create a mention candidate using the top element on the stack and the current buffer element. Score the candidate against existing clusters and resolve it (or create a new cluster).
\end{itemize}

The \peek\ action does not alter the stack but is otherwise identical to \pop.
This action is critical for detecting mentions sharing a left boundary.

\begin{figure}[t]
\begin{align*}
    \text{Initial} & \quad[\varnothing, 0, \varnothing, \varnothing] \\
    \text{Final} & \quad[\varnothing, n, A, \mathcal{C}] \\\\
    \push & \quad\frac{[S, i, A, \mathcal{C}]}{[S|w_i, i, A|\textsc{Push}, \mathcal{C}]} \\\\
    \adv & \quad\frac{[S, i, A, \mathcal{C}]}{[S, i+1, A|\textsc{Advance}, \mathcal{C}]} \\\\
    \pop\quad & \frac{[S|v, i, A, \mathcal{C}]}{[S, i, A|\pop, \mathcal{C}|\coref(v, w_i)]} \\\\
    \peek\quad & \frac{[S|v, i, A, \mathcal{C}]}{[S|v, i, A|\peek, \mathcal{C} | \coref(v, w_i)]}
\end{align*}
\caption{Deduction rules for our coreference resolver. $[S, i, A, \mathcal{C}]$ denotes the stack $S$, buffer index $i$, action history $A$, and cluster set $\mathcal{C}$. The \coref\ function indicates that span $(v, w_i)$ is clustered and added to $\mathcal{C}$.}
\label{fig:rules}
\end{figure}


Several hard action constraints ensure that only valid actions are taken and the final state is always reached. 
For example, \push\ can only be called once per token, or else the model would be marking the left boundary multiple times.
The full list of constraints is described in the appendix.

We denote the set of valid actions as $\mathcal{V}(S, i, A, \mathcal{C})$.
The conditional probability of selecting action $a_t$ based on state $\mathbf{p}_t$ can then be expressed as:

\[
p_M(a_t \vert\ \mathbf{p}_t) = \frac{\exp({w_{a_t} \cdot f_M (\mathbf{p}_t) )}}{\sum_{a'\in \mathcal{V}(S,i,A, \mathcal{C})}\exp{(w_{a'} \cdot f_M (\mathbf{p}_t))}},
\]

\noindent where $f_M$ is a two-layer neural network, and $w_{a_t}$ is a column vector selecting action $a_t$.

If \pop\ or \peek\ operations are predicted, the mention candidate is then scored against existing clusters.
Depending on these scores, the mention is either (a) resolved to an existing cluster, or (b) initialized as a new entity cluster. 
Define the set of possible coreference actions as $\mathcal{A}_k$, which includes resolving to existing clusters $\{\mathcal{C}_1,\dots,\mathcal{C}_k\}$ and creating new cluster $\mathcal{C}_{k+1}$.
We can write the conditional probability of coreference prediction $z_j$ based on mention candidate $m_j$ as:

\[
p_C(z_j \vert\ m_j) = \frac{\exp(w_{z_j}\cdot s_C(m_j))}{\sum_{z'\in \mathcal{A}_k } \exp(w_{z'}\cdot s_C(m_j))},
\]

\noindent where $s_C$ is a function scoring the mention candidate against $\{\mathcal{C}_1,\dots,\mathcal{C}_k, \mathcal{C}_{k+1}\}$ (described in Section \ref{sec:mc}).

The terminal state is reached when the final buffer element has been processed and the stack is empty.
At this point, all mentions have been clustered and we return all non-singleton entity clusters.
Figure \ref{fig:rules} presents a more formal description of the deduction rules, while an example is also shown in Figure \ref{fig:example}.

\begin{figure*}[t]
\centering
\begin{tabular}{l|l|l|l|c }
\toprule
Step & Action(s) & Stack & Buffer & Clusters \\\midrule
1. & \push & $[\text{\small Auto} ]$ & \small Auto workers ended their strike & $\varnothing$ \\
2. & \adv & $[\text{\small Auto} ]$ & \small workers ended their strike & $\varnothing$ \\
3. & \pop & $[\varnothing]$ & \small workers ended their strike & \{ \small Auto workers \}  \\
4. & \adv & $[\varnothing]$ & \small ended their strike & \{ \small Auto workers \} \\
5. & \adv & $[\varnothing]$ & \small their strike & \{ \small Auto workers \} \\
6. & \push & $[\text{\small their} ]$ & \small their strike & \{ \small Auto workers \} \\
7. & \peek & $[\text{\small their} ]$ & \small their strike & \{ \small Auto workers, their \} \\
8. & \adv & $[\text{\small their} ]$ & \small strike & \{ \small Auto workers, their \} \\
9. & \pop & $[\varnothing]$ & \small strike &  \{ \small Auto workers, their \} \{ their strike \} \\
10. & \adv & $[\varnothing]$ &  $\varnothing$ & \{ \small Auto workers, their \} \{ their strike \} \\
\bottomrule
\end{tabular}
\caption{Example of the shift-reduce system for the sentence ``Auto workers ended their strike''. $\varnothing$ denotes the empty stack or empty cluster set. Expressions within brackets mean they are co-referring. In each step, the Stack and Buffer show the result of applying the given action.}
\label{fig:example}
\end{figure*}

\subsection{Neural Implementation}
\subsubsection{Mention Detector}
Document tokens are first encoded using a pre-trained language model. 
The concatenated word embeddings, $x_1,\dots,x_n$, form the \textbf{buffer} for the shift-reduce mechanism.
Assuming current word $x_i$ and time step $t$, we denote the buffer as $\mathbf{b}_t = x_i$.

The stack is represented using a Stack-LSTM \cite{dyer-etal-2015-transition}. Let $x_{s_1},\dots,x_{s_L}$ be the currently marked left mention boundaries pushed to the stack. Then the \textbf{stack} representation at time $t$ is:
\[
    \mathbf{s}_t = \stacklstm[x_{s_1},\dots,x_{s_L}].
\]

We encode the action history $a_0,\dots,a_{t-1}$ with learned embeddings for each of the four actions. The action history at $t$ is encoded with an LSTM over previous action embeddings:
\[
    \mathbf{a}_t = \lstm[a_0,\dots,a_{t-1}]. 
\]

Then, the parser state is represented by the concatenation of buffer, stack, action history and additional mention features $\phi_M$:
\[
    \mathbf{p}_t = [\mathbf{b}_t; \mathbf{s}_t; \mathbf{a}_t; \phi_{M}(\mathbf{b}_t, \mathbf{s}_t)],
\]

\noindent where $\phi_M$ denotes learnable embeddings corresponding to useful mention features such as span width and document genre. For span width, we use embeddings measuring the distance from the top of the stack to the current buffer token (i.e. $i - s_L$), or $0$ if the stack is empty.

\subsubsection{Mention Clustering Model}\label{sec:mc}
The mention clustering is similar to previous online clustering methods \cite{toshniwal-etal-2020-learning, xia-etal-2020-incremental, webster-curran-2014-limited}, though we take care to avoid dependence on non-incremental pre-trained language models which have already been fine-tuned to this task.

Given a mention candidate's span representation $v$, we score $v$ against the existing entity cluster representations $\mathbf{m}_1,\dots,\mathbf{m}_k$:

\begin{align*}
    s_C(v) &= \left[ f_C(\mathbf{m}_1, v),\dots, f_C(\mathbf{m}_k, v), \alpha \right]\\
    f_C(\mathbf{m}_i, v) &= \mlp([v, \mathbf{m}_i, v\odot\mathbf{m}_i, \phi_C(v, \mathbf{m}_i)]) \\
    i^* &= \argmax_{i \in \{1,\dots,k+1 \}} s_C(v)
\end{align*}

\noindent where $f_C$ is two-layer neural network, $\alpha$ is a threshold value for creating a new cluster, $v\odot\mathbf{m}_i$ is the element-wise product and $\phi_C$ encodes useful features between $v$ and $\mathbf{m}_i$: the number of entities in $\mathbf{m}_i$, mention distance between $v$ and $\mathbf{m}_i$, the previous coreference action and document genre.

If the scores between $v$ and all cluster representations $\mathbf{m}_1,\dots,\mathbf{m}_k$ are below some threshold value $\alpha$ (i.e. $i^* = k+1$), we initialize a new entity cluster with $v$. Otherwise, we update the cluster representation $\mtop$ via a weighted average using the number of entities represented by $\mtop$:

\[
    \mtop \leftarrow \beta\cdot\mtop + (1-\beta)\cdot v,
\]
\noindent where $\beta = \frac{|\mtop|}{|\mtop| + 1}$ is the weighting term.


\subsubsection{Training}
Training is done via teaching forcing. At each time step, the model predicts the gold action given the present state.
The state is then updated using the gold action.
At each step, we compute mention detection loss $\mathcal{L}_M$ and coreference loss $\mathcal{L}_C$.

The mention detection loss $\mathcal{L}_M$ is calculated using the cross-entropy between the predicted mention detection action and gold action $a_{t^*} \in \mathcal{V}(S,i,A,\mathcal{C})$:
\[
    \mathcal{L}_M = - \sum_t \log p_M(a_{t^*}\ \vert\ \mathbf{p}_t),
\]
\noindent where $t$ sums over time steps across all documents.

Similarly, the coreference loss $\mathcal{L}_C$ is defined by the cross entropy between the highest-scoring coreference action and the gold coreference decision $z_{j^*} \in \mathcal{A}_k$:

\[
    \mathcal{L}_C = - \sum_j \log p_C(z_{j^*}\ \vert\ m_j),
\]

\noindent where $j$ sums over mentions across all documents. 
The entire network is then trained to optimize the sum of the two losses, $\mathcal{L}_M+\mathcal{L}_C$. During inference, we predict actions using greedy decoding, updating the state solely with predicted actions.

\begin{figure*}[t]
    \centering
    \begin{tikzpicture}
\node[inner sep=0pt] (doc) at (0,0) {\includegraphics[width=.10\textwidth]{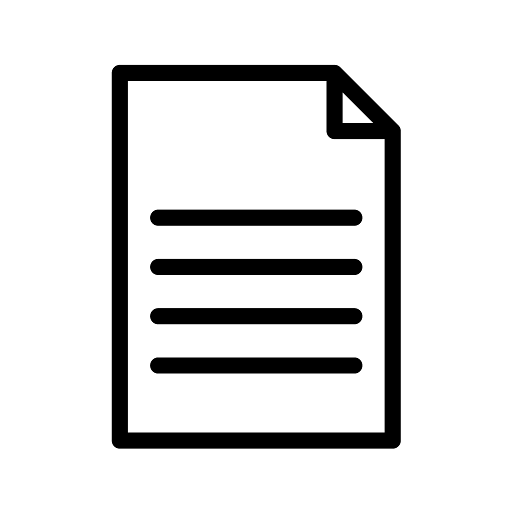}};

\node (encoder) [component, label={[align=center, yshift=0.3cm]Document\\Encoder}, right of=doc, xshift=1 cm] {};
\node (men_det) [component, label={[align=center, yshift=0.3cm]Mention\\Detector}, right of=encoder, xshift=1 cm, fill=greenx!30] {};
\node (coref) [component, label={[align=center, yshift=0.3cm]Coreference\\Resolver}, right of=men_det, xshift=1 cm, fill=purplex!30] {};

\matrix (mem) [matrix of nodes,thick,row sep=-\pgflinewidth,column sep=-\pgflinewidth, right of=coref, xshift=2cm, 
column 1/.style={nodes={rectangle,draw=black, anchor=base west,text width=2.5cm}},
label={[align=center, yshift=0.3cm]Cluster\\Embeddings}]  
{ \tiny $(s_1, s_2), (s_3, s_4)$ \\ \tiny $(s_5, s_6),\dots$ \\\tiny $(s_A, s_B)$ \\};

\draw [arrow] (doc) -- (encoder);
\draw [arrow] (encoder) -- node[anchor=south] {\tiny $[0.3,\dots]$} (men_det);
\draw [arrow] (men_det) -- node[anchor=south] {\tiny $(s_A,s_B)$} (coref) coordinate[midway] (men_mid);
\draw [arrow] (coref) -- (mem-1-1.west);
\draw [arrow] (coref) -- (mem-2-1.west) coordinate[midway] (coref_mid);
\draw [arrow] (coref) -- (mem-3-1.west);

\node (m_p) [below of=men_mid,yshift=-1.5cm] {\small $p_M(a_t \vert\ \mathbf{p}_t)$};
\draw [arrow, shorten >=0.4cm] (m_p) -- (men_mid);

\node (c_p) [below of=coref_mid,yshift=-1.5cm,xshift=0.5cm] {\small $p_C(z_t \vert\ (s_A,s_B))$};
\draw [arrow, shorten >=0.4cm] ([xshift=-0.5cm]c_p.north) -- (coref_mid);

\end{tikzpicture}
    \caption{A summary of the overall algorithm. 
    After document encoding, the mention detector predicts transition actions \push, \pop, \peek\ or \adv\ using the parser state $\mathbf{p}_t$.
    If a mention is predicted, the coreference resolver then clusters it to an existing cluster representation or creates a new cluster.
    Clustering a mention implies a coreference relation with mentions in the cluster.
    The steps can all be performed incrementally, assuming the document encoder is also incremental.
    }
    \label{fig:model-graph}
\end{figure*}
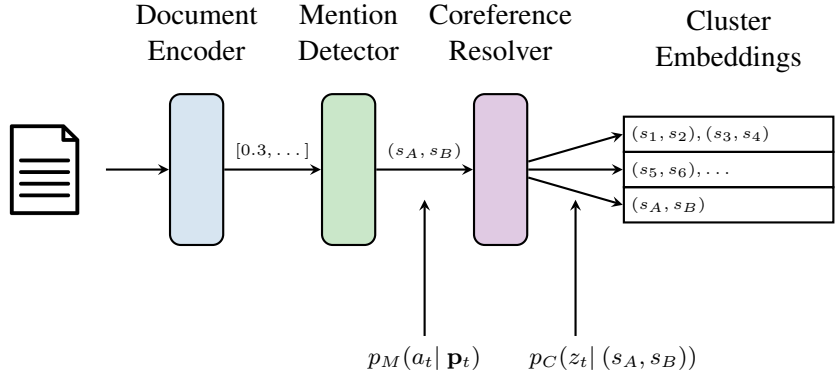

Figure \ref{fig:model-graph} presents a summary of the various components and the overall algorithm.

\section{Experiments}\label{sec:experiments}

\subsection{Datasets}\label{sec:datasets}
We train and evaluate our system on the \textbf{OntoNotes 5.0} dataset \cite{ontonotes}, using the same setup described in the CoNLL-2012 Shared Task \cite{pradhan-etal-2012-conll}.
OntoNotes includes 7 document genres and does not restrict mention token length; annotations cover pronouns, noun phrases and heads of verb phrases.
We evaluate using the \muc\ \cite{vilain-etal-1995-model}, \bcub\  \cite{bagga1998algorithms} and \ceaf\ \cite{luo-2005-coreference} metrics and their average (the CoNLL score), using the official CoNLL-2012 scorer.

We also test models on the the recently released \textbf{CODI-CRAC 2021} corpus \cite{khosla-etal-2021-codi}.
This dataset annotates coreference (and other anaphora-related tasks) for 134 documents across 4 separate dialogue corpora (Light, \citealt{urbanek-etal-2019-learning}, AMI, \citealt{ami}, Persuasion, \citealt{wang-etal-2019-persuasion} and Switchboard,  \citealt{switchboard}).
The dataset suits incremental systems well since dialogue can be naturally presented as incremental utterances.
Given the small dataset size, we use it for evaluation only, using models trained on OntoNotes.
Since OntoNotes marks document genre (which systems often use as a feature), we associate CODI-CRAC documents with OntoNotes' `telephone conversation' genre, since it is the most similar.
We remove singleton clusters due to lack of annotation in the training set.
We again evaluate using \muc, \bcub\ and \ceaf, using the official Universal Anaphora scorer \cite{yu-etal-2022-universal}.

\subsection{Model Components}
\subsubsection{Document Encoder}
Recent models on coreference resolution often use SpanBERT \cite{joshi-etal-2020-spanbert} for word embeddings (\citet{wu-etal-2020-corefqa, toshniwal-etal-2020-learning, xia-etal-2020-incremental, xu-choi-2020-revealing}, among others), since  \citet{joshi-etal-2020-spanbert} demonstrate SpanBERT's proficiency for entity-related tasks such as coreference resolution.
However, SpanBERT is unsuitable for incremental applications because it expects all its input simultaneously and cannot partially process text while waiting for future input.

Instead, we turn to XLNet\footnote{We use the \textit{base} version due to memory restrictions.} \cite{xlnet}, which extends the earlier Transformer-XL \cite{dai-etal-2019-transformer}.
XLNet differs from typical pre-trained language models as it can efficiently cache and reuse its previous outputs. 
The caching mechanism allows for recurrent computation to be performed efficiently.
Cached outputs provide a context to the current sentence being processed.

We experiment using XLNet in two settings: in the \textbf{Sentence-Incremental} (Sent-Inc) setting, each sentence is processed sequentially, and partial coreference clusters are computed before the next sentence is observed. 
After each sentence is processed, we accumulate XLNet's outputs (up to a cutoff point) and reuse them when processing the next sentence.
We limit the number of cached tokens so that the cached and `active' tokens do not exceed 512, so that our work remains comparable to other recent works.
Although the mention detector is token-incremental and the mention clustering component is span-incremental, the document encoder is sentence-incremental, so overall we describe the system as sentence-incremental.

In the \textbf{Part-Incremental} (Part-Inc) setting, we allow XLNet to access multiple sentences simultaneously.
We experiment both with and without the cache mechanism, using up to a total of 512 tokens at a time.
This setting is comparable to experiments in \citet{xia-etal-2020-incremental} and \citet{toshniwal-etal-2020-learning}, where document encoding is also non-incremental.
In our case, both mention detection and mention clustering components remain incremental as in the Sentence-Incremental setting.
In this way, we can isolate the effect of sentence-incrementality on the document encoder (XLNet).


\subsubsection{Span Representation}
We use a similar span representation to \citet{lee-etal-2017-end}: for a span $(i, j)$, we concatenate word embeddings $(x_i, x_j)$, an attention-weighted average $\bar{x}$ and learnable embeddings for span width and speaker ID (the speaker for $(i, j)$). We use 20-dimensional learned embeddings for all features (span width, speaker ID, document genre, action history, mention distance and number of entities in each cluster).

\subsubsection{Training}
We use Adam to train task-specific parameters, and AdamW for XLNet's parameters \cite{KingmaB14, loshchilov2018decoupled}.
The gradient is accumulated across one document before updating model weights.
We use a learning rate scheduler with a linear decay, and additionally warmup SpanBERT's parameters for the first 10\% update steps.
For the mention detector, we balance the loss weights based on the frequency of each action in the training set. 
This step is important because most tokens do not correspond to mention boundaries, meaning the \adv\ action is by far the most prevalent in the training set.

Training converges within 15 epochs. The model is implemented in PyTorch \cite{pytorch}. 
A complete list of hyperparameters is included in the appendix.

\subsection{Comparisons}

We compare against several recent works with varying degrees of incrementality. 
Table \ref{tab:compare} summarizes their differences in incrementality compared to ours, as well as the span complexity.
\citet{joshi-etal-2020-spanbert} is a non-incremental formulation: it adopts the \textit{end-to-end} and \textit{coarse-to-fine} formulations from \citet{lee-etal-2017-end} and \citet{lee-etal-2018-higher}, replacing the LSTM encoder with their novel SpanBERT architecture.

\begin{table}[h]
\centering
\small
\begin{tabular}{l l l}
\toprule					
Model	&	Incremental	&	Span	\\
	&	Components	&	Complexity	\\
\midrule					
SpanBERT	&	None	&	$O(n^4)$	\\
\midrule					
longdoc	&	Mention Clustering	&	$O(n^2m)$	\\
ICoref	&	Mention Clustering	&	$O(n^2m)$	\\
\midrule					
Part-Inc (Ours)	&	Mention Detection +	&	$O(nm)$	\\
	&	Mention Clustering	&		\\
\midrule					
ICoref-\textit{inc}	&	All	&	$O(n^2m)$	\\
Sent-Inc (Ours)	&	All	&	$O(nm)$	\\
\bottomrule	
\end{tabular}
\caption{The list of systems we compare, alongside their incrementality (on a sentence-level) and span complexity.
`All Components' means document encoding, mention detection and mention clustering.
$n$ is the number of tokens and $m$ is the number of entities.}
\label{tab:compare}
\end{table}

The longdoc \cite{toshniwal-etal-2020-learning} and ICoref \cite{xia-etal-2020-incremental} systems adapt \citet{joshi-etal-2020-spanbert} so that mention clustering is done incrementally. 
However, both models avoid modifying the non-incremental document encoding and mention detection steps from \citet{joshi-etal-2020-spanbert}, and the resulting systems are only partly incremental.
Since \citet{toshniwal-etal-2020-learning} and \citet{xia-etal-2020-incremental} only experiment with SpanBERT-large, we re-train their implementations with SpanBERT-base to fairly compare against our own systems.

\citet{xia-etal-2020-incremental} also provide a truly sentence-incremental version of their system, which we call ICoref-\textit{inc}\footnote{Specifically, this system is the ``Train 1-sentence / Inference 1-sentence'' model from \citet{xia-etal-2020-incremental}'s Table 4.}.
This version is trained by encoding tokens and proposing mentions sentence-by-sentence, independently processing each sentence as it is observed while maintaining entity clusters across sentences.
Since ICoref-\textit{inc} is fully sentence-incremental, it provides the fairest comparison to our own Sentence-Incremental setting.
Having more incremental components results in increased difficulty on the coreference task, as the system must rely on partial information when making clustering decisions.

We do not compare against \citet{liu-etal-2019-referential} and \citet{toshniwal-etal-2020-petra}'s token-incremental models.
Besides being generally unsuitable for span-based coreference, they also do not handle nested mentions.
Roughly 11\% of OntoNotes' mentions are nested, meaning that training these systems on OntoNotes is infeasible.

\subsubsection{Span Complexity}
Table \ref{tab:compare} also compares the \textit{span complexity} between systems, in terms of how many spans must be scored and compared.
This comparison is analytic and not runtime-based, and so ignores handcrafted memory-saving techniques such as eviction and span pruning.
\citet{joshi-etal-2020-spanbert} score all possible spans and compare them pairwise, meaning their system runs in $O(n^4)$, where $n$ is the number of tokens.
\citet{toshniwal-etal-2020-learning} and \citet{xia-etal-2020-incremental} reduce the complexity to $O(n^2m)$, where $m$ is the number of entities, by incrementally clustering mentions.
Finally, we claim our systems' span complexity is $O(nm)$.
Our mention detector proposes $O(n)$ spans, as we can show each action is linearly bounded in the number of tokens. 
Our reduced complexity speaks to its increased cognitive plausibility compared to the part- and non-incremental systems, which consider a quadratic number of spans.

Note that the runtime is not comparable because non-incremental methods process the entire document in parallel, whereas ours is not parallelizable and therefore slower.
We also note that this comparison does not have any bearing on memory requirements, since \citet{toshniwal-etal-2020-learning} and \citet{xia-etal-2020-incremental} both maintain constant memory through eviction strategies.

\begin{table*}[t]
\small
\centering
\begin{tabular}{l|c|c c c c c c c c c|c}
\toprule																							
	&		&	\multicolumn{3}{c}{\muc}					&	\multicolumn{3}{c}{\bcub}					&	\multicolumn{3}{c}{\ceaf}					&	Avg.	\\
Model	&	SI	&	Rec.	&	Prec.	&	F1	&	Rec.	&	Prec.	&	F1	&	Rec.	&	Prec.	&	F1	&	F1	\\
\midrule																							
SpanBERT-l \cite{joshi-etal-2020-spanbert}	&	None	&	84.8	&	85.8	&	85.3	&	77.9	&	78.3	&	78.1	&	74.2	&	76.4	&	75.3	&	79.6	\\
CorefQA-l \cite{wu-etal-2020-corefqa}	&	None	&	87.4	&	88.6	&	88.0	&	82.0	&	82.4	&	82.2	&	78.3	&	79.9	&	79.1	&	83.1	\\
s2e  \cite{kirstain-etal-2021-coreference}	&	None	&	85.1	&	86.5	&	85.8	&	77.9	&	80.3	&	79.1	&	75.4	&	76.8	&	76.1	&	80.3	\\
s2e+se\_ct\cite{chai-strube-2022-incorporating}	&	None	&	85.3	&	87.2	&	86.3	&	78.6	&	80.7	&	79.6	&	75.2	&	78.2	&	76.7	&	80.9	\\
\midrule																							
SpanBERT-b \cite{joshi-etal-2020-spanbert}	&	None	&	83.1	&	84.3	&	83.7	&	75.3	&	76.2	&	75.8	&	71.2	&	74.6	&	72.9	&	77.4	\\
CorefQA-b \cite{wu-etal-2020-corefqa}	&	None	&	87.4	&	85.2	&	86.3	&	76.5	&	78.7	&	77.6	&	75.6	&	76.0	&	75.8	&	79.9	\\
\midrule																							
longdoc \cite{toshniwal-etal-2020-learning}	&	Part	&	83.3	&	83.0	&	83.2	&	75.5	&	74.1	&	74.8	&	70.1	&	72.8	&	71.4	&	76.4	\\
ICoref \cite{xia-etal-2020-incremental}	&	Part	&	83.1	&	84.2	&	83.6	&	74.3	&	75.8	&	75.0	&	71.7	&	73.3	&	72.5	&	77.0	\\
Part-Inc (Ours)	&	Part	&	83.7	&	82.1	&	82.9	&	75.9	&	73.0	&	74.4	&	68.8	&	74.5	&	71.6	&	76.3	\\
\midrule																							
ICoref-\textit{inc} \cite{xia-etal-2020-incremental}	&	All	&	74.0	&	79.7	&	76.7	&	58.6	&	70.6	&	64.0	&	63.7	&	63.1	&	63.4	&	68.0	\\
Sent-Inc (Ours)	&	All	&	78.1	&	79.4	&	78.8	&	68.9	&	68.3	&	68.6	&	55.8	&	71.2	&	62.5	&	70.0	\\
\bottomrule				
\end{tabular}
\caption{Main results on the OntoNotes 5.0 test set with the CoNLL 2012 Shared Task metrics and the average F1 (the CoNLL F1 score). 
The `SI' column denotes the sentence-incrementality of each system, summarizing details in Table \ref{tab:compare}.
The top four systems are not directly comparable to ours, since they train with a `large' encoder (either SpanBERT or Longformer \cite{Beltagy2020Longformer}).
Note that scores for \citet{xia-etal-2020-incremental} and \citet{toshniwal-etal-2020-learning} differ from their reported results because we re-train them with SpanBERT-\textit{base} instead of \textit{large}.
}
\label{tab:results}
\end{table*}

\begin{table}[t]
    \small
    \centering
    \begin{tabular}{l|c c c c|c}
\toprule											
Model	&	\textsc{Light}	&	\textsc{AMI}	&	\textsc{Pers.}	&	\textsc{Swbd.}	&	Avg.	\\
\midrule											
SpanB	&	57.7	&	33.8	&	53.7	&	50.2	&	48.9	\\
\midrule											
ICoref	&	54.7	&	33.7	&	51.5	&	48.1	&	47.0	\\
Part-Inc	&	53.5	&	32.4	&	50.5	&	46.9	&	45.8	\\
\midrule											
IC-\textit{inc}	&	45.5	&	21.9	&	41.3	&	36.6	&	36.3	\\
Sent-Inc	&	50.5	&	31.6	&	46.4	&	44.7	&	43.3	\\
\bottomrule			
    \end{tabular}
    \caption{Main results for the CODI-CRAC 2021 corpus. All scores denote the CoNLL F1 score (average of \muc, \bcub\ and \ceaf). Here, SpanB=SpanBERT, and IC-\textit{inc}=ICoref-\textit{inc}.}
    \label{tab:codi-crac-res}
\end{table}

\section{Results}
\subsection{OntoNotes}
The main results for OntoNotes are shown in Table \ref{tab:results}. 
First, SpanBERT \cite{joshi-etal-2020-spanbert}, being non-incremental, unsurprisingly outperforms other systems, both part and sentence incremental.

Within partly incremental systems, the ICoref model \cite{xia-etal-2020-incremental} performs best, below SpanBERT by 0.4 F1. 
Our Part-Inc model performs comparably to longdoc \cite{toshniwal-etal-2020-learning}, only trailing ICoref by 0.7 F1 points.


The advantages of our method are more evident in the sentence-incremental evaluation.
Since ICoref-\textit{inc} relies on SpanBERT to encode tokens and score mentions, its performance suffers considerably when evaluated in the sentence-incremental setting.
In contrast, the Sent-Inc model effectively uses the history of previous processed sentences and outperforms ICoref-\textit{inc} by 2 F1 points. 
Still, both systems suffer considerably when compared to their part-incremental counterparts: ICoref drops by 9 F1 points and our model by 6.3 F1.
In Section \ref{sec:analysis}, we explore the main causes of this drop.



\subsection{CODI-CRAC}
The results on the CODI-CRAC corpus are shown in Table \ref{tab:codi-crac-res}. We observe many of the same trends as in OntoNotes: the non-incremental SpanBERT again surpasses other models, achieving 2.9 F1 higher than ICoref.

Within partly incremental systems, our Part-Inc system trails ICoref by 1.2 F1.
We omit the longdoc results from this table, after finding its performance surprisingly plummets when evaluated on CODI-CRAC.
On all subsets, it scores below 2 F1, indicating issues with model transfer.
Other works have explored this topic in depth \cite{toshniwal-etal-2021-generalization}, and we do not investigate further here.

In the Sentence Incremental setting, although our Sent-Inc model again outperforms \citet{xia-etal-2020-incremental}'s ICoref-\textit{inc}, the performance difference is much larger here: 7 F1 compared to 2 F1 in OntoNotes.
The gap between the Sent-Inc and Part-Inc is also much smaller: only 2.5 F1 points compared to 6.3 F1 on OntoNotes.
The difference in performances between the two datasets may suggest our model is better suited to the inherent incrementality in a dialogue setting.

\section{Analysis}\label{sec:analysis}
The dramatic performance gap between the Sent-Inc and Part-Inc settings may be surprising.
Since coreference resolution is primarily processed incrementally by humans, why does access to future tokens affect the Sent-Inc model so heavily?

To investigate this issue deeper, we design additional $k$-Sentence-Incremental settings.
In each setting, the system accesses $k$ sentences ($S_1,\dots,S_k$) at a time as active input, and $512 - \sum_{i=1}^{k}|S_k|$ tokens as memory.
In each setting, the model observes the same number of tokens (512), but varies the amount of active input vs. memory.
The mention detection and mention clustering steps remain the same and are still incremental; the only change is in the encoder.

Varying $k$ in this way allows us to the test the effect of more or less incrementality on the system.
When $k = 1$, we recover the original Sent-Inc model.
When $k$ is large enough (in practice, 24), we get the Part-Inc model.
For each $k \in \{1,4,8,12,16,20,24\}$, we fully train the corresponding model on OntoNotes as described in Section \ref{sec:experiments}, and evaluate on the dev set.

\begin{figure}[ht]
    \centering
    \resizebox{\columnwidth}{!}{
    \begin{tikzpicture}
\definecolor{darkslategray38}{RGB}{38,38,38}
\definecolor{lavender234234242}{RGB}{234,234,242}
\definecolor{lightgray204}{RGB}{204,204,204}
\definecolor{mediumseagreen85168104}{RGB}{85,168,104}
\definecolor{peru22113282}{RGB}{221,132,82}
\definecolor{steelblue76114176}{RGB}{76,114,176}

\begin{axis}[
axis background/.style={fill=lavender234234242},
axis line style={white},
legend cell align={left},
legend style={
  fill opacity=0.8,
  draw opacity=1,
  text opacity=1,
  at={(0.97,0.03)},
  anchor=south east,
  draw=lightgray204,
  fill=lavender234234242
},
legend entries={Recall, Precision, F1},
tick align=outside,
x grid style={white},
xlabel=\textcolor{darkslategray38}{\(\displaystyle k\)-Sentence Setting},
xmajorgrids,
xmajorticks=true,
xtick pos=bottom,
xmin=-0.15, xmax=25.15,
xtick style={color=darkslategray38},
y grid style={white},
ylabel=\textcolor{darkslategray38}{Score},
ymajorgrids,
ymajorticks=true,
ymin=66.29, ymax=77.51,
ytick style={color=darkslategray38},
ytick pos=left
]
\addplot [very thick, steelblue76114176]
table[row sep=crcr] {%
1 66.8 \\
4 72.7 \\
8 73.9 \\
12 74.4 \\
16 76.2 \\
20 76.2 \\
24 76.5 \\
};
\addplot [very thick, peru22113282]
table[row sep=crcr] {%
1 72.5 \\
4 74.7 \\
8 75.4 \\
12 76.4 \\
16 75.8 \\
20 77 \\
24 76.3 \\
};
\addplot [very thick, mediumseagreen85168104]
table[row sep=crcr] {%
1 69.4 \\
4 73.6 \\
8 74.6 \\
12 75.9 \\
16 75.9 \\
20 76.6 \\
24 76.4 \\
};
\end{axis}

\end{tikzpicture}
    }
    \caption{The CoNLL performance (average of \muc, \bcub\ and \ceaf) of each $k$-Sentence-Incremental model on the OntoNotes dev set.}
    \label{fig:kscores}
\end{figure}
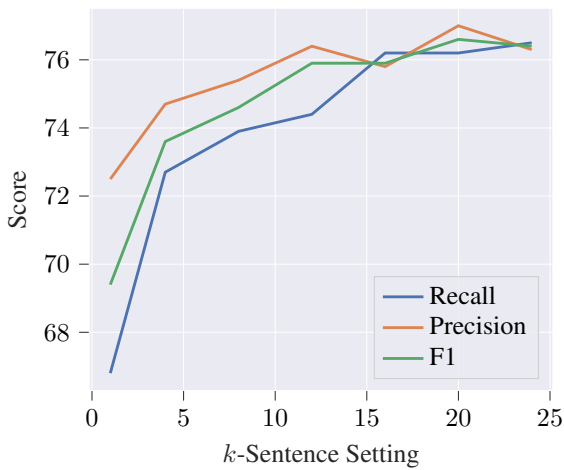

The results are shown in Figure \ref{fig:kscores}.
There are a few notable characteristics.
The first is that as $k$ increases, we see a much more dramatic lift when $k$ is small (e.g. moving from 1 to 4 sentences) compared to when $k$ is large.
This effect corresponds to the intuition that coreferring expressions are usually close to their antecedent.
The more coreference chains the model can observe simultaneously, the better it is at resolving them.

The second noteworthy trend is that increasing $k$ improves recall (9.7\%) far more than precision (3.8\%).
Although not shown here, we observe this trend across all three metrics within the CoNLL score (\muc,\ \bcub\ and \ceaf).
The result means that finding and resolving true coreference links (i.e. reducing false negatives) is a far more serious obstacle for the Sent-Inc model 
than for Part-Inc.
Since the only difference in these models is how many embeddings are cached, the result suggests caching or not caching embeddings plays a large role in finding and correctly resolving mentions.



\section{Future Work}
A major goal would be to elevate incremental coreference resolvers to the same level as non-incremental ones.
As we showed in Section \ref{sec:analysis}, a large part of the performance difference occurs because the XLNet encoder does not effectively handle incremental input.
A simple strategy therefore may be to swap out that encoder for a more powerful one.
However, few pre-trained language models targeted at NLU tasks are naturally incremental. 
One candidate is GPT-J \cite{gpt-j} but its size is prohibitively large.

Other ways to bridge this gap may come from improving the mention detection component. 
A similar task is nested named entity recognition, where the system must identify named entity boundaries and coarsely classify them. 
Recent nested NER systems such as \citet{katiyar-cardie-2018-nested} or \citet{yu-etal-2020-named} may provide directions for improving mention detection in our incremental formulation.



\section{Conclusion}
We propose a sentence-incremental coreference resolution model using a shift-reduce formulation.
The model delays mention clustering until the full span has been observed, alleviating a key flaw with previous incremental systems.
It efficiently processes text, and avoids scoring a quadratic number of spans during mention detection.

In a sentence-incremental setting, our method outperforms strong baselines adapted from state-of-the-art systems.
When access to the full document is allowed, the proposed system achieves similar performance to state-of-the-art methods while maintaining a higher level of incrementality.
We investigate why this relaxation has such a dramatic effect, finding that the document encoder does not make effective use of its memory cache.

Our sentence-incremental results suggest an important point: non-incremental methods are not effective tools when they must be used incrementally.
Creating new, incremental coreference resolvers that perform at the same level as non-incremental ones is a challenging but meaningful goal.
Achieving this result would make a significant impact in downstream applications where text is received incrementally, such as dialogue systems or conversational question answering (e.g. \citealt{andreas-etal-2020-task, martin-etal-2020-mudoco}).
Our proposal demonstrates an important step towards highly effective, incremental coreference resolution systems.

\section*{Limitations}
In this work, we have experimented with training neural networks on OntoNotes and evaluating on other datasets (the CODI-CRAC 2021 corpus).
Several recently published papers have explored the difficulties of coreference resolution model transfer \cite{subramanian-roth-2019-improving,xia-van-durme-2021-moving,toshniwal-etal-2021-generalization, yuan-etal-2022-adapting}.
These works have noted generalization problems with models trained on OntoNotes, with one particular difficulty that OntoNotes does not annotate singleton clusters, or `markable' mentions.

Several recent works have addressed generalization issues by training on additional resources \cite{subramanian-roth-2019-improving,xia-van-durme-2021-moving,toshniwal-etal-2021-generalization, yuan-etal-2022-adapting}.
In particular, \citet{toshniwal-etal-2021-generalization} augment OntoNotes with \textit{pseudo-singletons}: a fully trained coreference resolver scores all spans in the text, and the top-scoring spans outside of gold mentions are regarded as singletons.
The authors show that adding pseudo-singletons to the OntoNotes training data improves (1) coreference resolution metrics on OntoNotes and (2) generalization capabilities.
Pseudo-singletons are especially helpful for transfer learning from OntoNotes because other coreference datasets will often annotate singletons.

In our experiments, we attempted to use their published pseudo-singletons, but faced difficulties because the pseudo-singletons do not respect the ``non-crossing bracketing'' structure in OntoNotes, and overlap arbitrarily (not only nested). 
Our mention detector assumes mentions may be nested but otherwise do not overlap, and determining which pseudo-singletons to filter out without redoing the whole experiment was not feasible. 
We leave the problem for future work, but we agree that models trained on OntoNotes without heuristically added singletons are limited in their generalization capabilities.

Our experiments have focused on the \textit{base} versions of XLNet and SpanBERT due to resource requirements.
Training our models requires a GPU with 16 GB of memory; we used NVIDIA Tesla V100 16 GB cards.
Greater memory efficiency could be achieved by extending the memory to be more dynamic.
In the current system, entities are added but never evicted.
Ideally, when a referent is no longer relevant to the context, it should be detected and removed.
This concept has been explored with memory network-based systems \cite{liu-etal-2019-referential,toshniwal-etal-2020-petra}, and also recent partly incremental systems \cite{xia-etal-2020-incremental,toshniwal-etal-2020-learning}.
Memory-based systems using dynamic eviction strategies appear in other NLP tasks as well, such as semantic parsing \cite{jain-memory-2021}.

\section*{Ethical Considerations}
NLP systems such as ours must be employed with special consideration that they do not demonstrate unwanted patterns towards protected groups.
Previously, systems have been shown to learn harmful associations from training corpora. 
For example, \citet{bolukbasi} show that word embeddings trained on a news corpus exhibit gender stereotypes, such as associating ``receptionist'' with ``female''.

Coreference resolution systems in particular may learn gender biases, and methods exist to counter this effect \cite{rudinger-etal-2018-gender, zhao-etal-2018-gender}.
Our system is trained on OntoNotes, which include data from a diverse set of sources such as Wall Street Journal articles, telephone conversations and Bible passages. 
Our final trained model may therefore reflect undesirable content from these texts.

Any off-the-shelf deployment of our model should first check whether the model is harmful towards any protected group, and appropriate mitigation should be taken.
For example, evaluating on specialized datasets such as \citet{webster-etal-2018-mind} may indicate whether the system unfairly predicts certain labels based on gender.

\section*{Acknowledgements}
We would like to thank Ratish Puduppully and Miloš Stanojević as well as the anonymous reviewers for their helpful discussions and feedback.
This work used the Cirrus UK National Tier-2 HPC Service at EPCC (\href{http://www.cirrus.ac.uk}{http://www.cirrus.ac.uk}) funded by the University of Edinburgh and EPSRC (EP/P020267/1), as well as the Baskerville Tier 2 HPC service (\href{https://www.baskerville.ac.uk/}{https://www.baskerville.ac.uk/}). Baskerville was funded by the EPSRC and UKRI through the World Class Labs scheme (EP/T022221/1) and the Digital Research Infrastructure programme (EP/W032244/1) and is operated by Advanced Research Computing at the University of Birmingham.
The work was partly supported by ERC Advanced Fellowship GA 742137 SEMANTAX and the University of Edinburgh Huawei Laboratory.

\bibliography{anthology,custom}

\begin{thebibliography}{56}
\expandafter\ifx\csname natexlab\endcsname\relax\def\natexlab#1{#1}\fi

\bibitem[{Allaway et~al.(2021)Allaway, Wang, and
  Ballesteros}]{allaway-etal-2021-sequential}
Emily Allaway, Shuai Wang, and Miguel Ballesteros. 2021.
\newblock \href {https://doi.org/10.18653/v1/2021.emnlp-main.382} {Sequential
  cross-document coreference resolution}.
\newblock In \emph{Proceedings of the 2021 Conference on Empirical Methods in
  Natural Language Processing}, pages 4659--4671, Online and Punta Cana,
  Dominican Republic. Association for Computational Linguistics.

\bibitem[{Altmann and Steedman(1988)}]{altmann}
Gerry Altmann and Mark Steedman. 1988.
\newblock \href {https://doi.org/https://doi.org/10.1016/0010-0277(88)90020-0}
  {Interaction with context during human sentence processing}.
\newblock \emph{Cognition}, 30(3):191--238.

\bibitem[{Andreas et~al.(2020)Andreas, Bufe, Burkett, Chen, Clausman, Crawford,
  Crim, DeLoach, Dorner, Eisner, Fang, Guo, Hall, Hayes, Hill, Ho, Iwaszuk,
  Jha, Klein, Krishnamurthy, Lanman, Liang, Lin, Lintsbakh, McGovern,
  Nisnevich, Pauls, Petters, Read, Roth, Roy, Rusak, Short, Slomin, Snyder,
  Striplin, Su, Tellman, Thomson, Vorobev, Witoszko, Wolfe, Wray, Zhang, and
  Zotov}]{andreas-etal-2020-task}
Jacob Andreas, John Bufe, David Burkett, Charles Chen, Josh Clausman, Jean
  Crawford, Kate Crim, Jordan DeLoach, Leah Dorner, Jason Eisner, Hao Fang,
  Alan Guo, David Hall, Kristin Hayes, Kellie Hill, Diana Ho, Wendy Iwaszuk,
  Smriti Jha, Dan Klein, Jayant Krishnamurthy, Theo Lanman, Percy Liang,
  Christopher~H. Lin, Ilya Lintsbakh, Andy McGovern, Aleksandr Nisnevich, Adam
  Pauls, Dmitrij Petters, Brent Read, Dan Roth, Subhro Roy, Jesse Rusak, Beth
  Short, Div Slomin, Ben Snyder, Stephon Striplin, Yu~Su, Zachary Tellman, Sam
  Thomson, Andrei Vorobev, Izabela Witoszko, Jason Wolfe, Abby Wray, Yuchen
  Zhang, and Alexander Zotov. 2020.
\newblock \href {https://doi.org/10.1162/tacl_a_00333} {Task-oriented dialogue
  as dataflow synthesis}.
\newblock \emph{Transactions of the Association for Computational Linguistics},
  8:556--571.

\bibitem[{Bagga and Baldwin(1998)}]{bagga1998algorithms}
Amit Bagga and Breck Baldwin. 1998.
\newblock Algorithms for scoring coreference chains.
\newblock In \emph{The first international conference on language resources and
  evaluation workshop on linguistics coreference}, volume~1, pages 563--566.
  Citeseer.

\bibitem[{Beltagy et~al.(2020)Beltagy, Peters, and
  Cohan}]{Beltagy2020Longformer}
Iz~Beltagy, Matthew~E. Peters, and Arman Cohan. 2020.
\newblock Longformer: The long-document transformer.
\newblock \emph{arXiv:2004.05150}.

\bibitem[{Bolukbasi et~al.(2016)Bolukbasi, Chang, Zou, Saligrama, and
  Kalai}]{bolukbasi}
Tolga Bolukbasi, Kai-Wei Chang, James Zou, Venkatesh Saligrama, and Adam Kalai.
  2016.
\newblock Man is to computer programmer as woman is to homemaker? debiasing
  word embeddings.
\newblock In \emph{Proceedings of the 30th International Conference on Neural
  Information Processing Systems}, NIPS'16, page 4356–4364, Red Hook, NY,
  USA. Curran Associates Inc.

\bibitem[{Carletta(2006)}]{ami}
J.~Carletta. 2006.
\newblock Announcing the ami meeting corpus.
\newblock In \emph{The ELRA Newsletter 11(1)}, pages 3 -- 5.

\bibitem[{Chai and Strube(2022)}]{chai-strube-2022-incorporating}
Haixia Chai and Michael Strube. 2022.
\newblock \href {https://doi.org/10.18653/v1/2022.naacl-main.218}
  {Incorporating centering theory into neural coreference resolution}.
\newblock In \emph{Proceedings of the 2022 Conference of the North American
  Chapter of the Association for Computational Linguistics: Human Language
  Technologies}, pages 2996--3002, Seattle, United States. Association for
  Computational Linguistics.

\bibitem[{Dai et~al.(2019)Dai, Yang, Yang, Carbonell, Le, and
  Salakhutdinov}]{dai-etal-2019-transformer}
Zihang Dai, Zhilin Yang, Yiming Yang, Jaime Carbonell, Quoc Le, and Ruslan
  Salakhutdinov. 2019.
\newblock \href {https://doi.org/10.18653/v1/P19-1285} {Transformer-{XL}:
  Attentive language models beyond a fixed-length context}.
\newblock In \emph{Proceedings of the 57th Annual Meeting of the Association
  for Computational Linguistics}, pages 2978--2988, Florence, Italy.
  Association for Computational Linguistics.

\bibitem[{Dyer et~al.(2015)Dyer, Ballesteros, Ling, Matthews, and
  Smith}]{dyer-etal-2015-transition}
Chris Dyer, Miguel Ballesteros, Wang Ling, Austin Matthews, and Noah~A. Smith.
  2015.
\newblock \href {https://doi.org/10.3115/v1/P15-1033} {Transition-based
  dependency parsing with stack long short-term memory}.
\newblock In \emph{Proceedings of the 53rd Annual Meeting of the Association
  for Computational Linguistics and the 7th International Joint Conference on
  Natural Language Processing (Volume 1: Long Papers)}, pages 334--343,
  Beijing, China. Association for Computational Linguistics.

\bibitem[{Godfrey et~al.(1992)Godfrey, Holliman, and McDaniel}]{switchboard}
John~J. Godfrey, Edward~C. Holliman, and Jane McDaniel. 1992.
\newblock Switchboard: Telephone speech corpus for research and development.
\newblock In \emph{Proceedings of the 1992 IEEE International Conference on
  Acoustics, Speech and Signal Processing - Volume 1}, ICASSP'92, page
  517–520, USA. IEEE Computer Society.

\bibitem[{Jain and Lapata(2021)}]{jain-memory-2021}
Parag Jain and Mirella Lapata. 2021.
\newblock \href {https://doi.org/10.1162/tacl_a_00422} {{Memory-Based Semantic
  Parsing}}.
\newblock \emph{Transactions of the Association for Computational Linguistics},
  9:1197--1212.

\bibitem[{Joshi et~al.(2020)Joshi, Chen, Liu, Weld, Zettlemoyer, and
  Levy}]{joshi-etal-2020-spanbert}
Mandar Joshi, Danqi Chen, Yinhan Liu, Daniel~S. Weld, Luke Zettlemoyer, and
  Omer Levy. 2020.
\newblock \href {https://doi.org/10.1162/tacl_a_00300} {{S}pan{BERT}: Improving
  pre-training by representing and predicting spans}.
\newblock \emph{Transactions of the Association for Computational Linguistics},
  8:64--77.

\bibitem[{Joshi et~al.(2019)Joshi, Levy, Zettlemoyer, and
  Weld}]{joshi-etal-2019-bert}
Mandar Joshi, Omer Levy, Luke Zettlemoyer, and Daniel Weld. 2019.
\newblock \href {https://doi.org/10.18653/v1/D19-1588} {{BERT} for coreference
  resolution: Baselines and analysis}.
\newblock In \emph{Proceedings of the 2019 Conference on Empirical Methods in
  Natural Language Processing and the 9th International Joint Conference on
  Natural Language Processing (EMNLP-IJCNLP)}, pages 5803--5808, Hong Kong,
  China. Association for Computational Linguistics.

\bibitem[{Kahardipraja et~al.(2021)Kahardipraja, Madureira, and
  Schlangen}]{kahardipraja-etal-2021-towards}
Patrick Kahardipraja, Brielen Madureira, and David Schlangen. 2021.
\newblock \href {https://doi.org/10.18653/v1/2021.emnlp-main.90} {Towards
  incremental transformers: An empirical analysis of transformer models for
  incremental {NLU}}.
\newblock In \emph{Proceedings of the 2021 Conference on Empirical Methods in
  Natural Language Processing}, pages 1178--1189, Online and Punta Cana,
  Dominican Republic. Association for Computational Linguistics.

\bibitem[{Kasai et~al.(2021)Kasai, Peng, Zhang, Yogatama, Ilharco, Pappas, Mao,
  Chen, and Smith}]{kasai-etal-2021-finetuning}
Jungo Kasai, Hao Peng, Yizhe Zhang, Dani Yogatama, Gabriel Ilharco, Nikolaos
  Pappas, Yi~Mao, Weizhu Chen, and Noah~A. Smith. 2021.
\newblock \href {https://doi.org/10.18653/v1/2021.emnlp-main.830} {Finetuning
  pretrained transformers into {RNN}s}.
\newblock In \emph{Proceedings of the 2021 Conference on Empirical Methods in
  Natural Language Processing}, pages 10630--10643, Online and Punta Cana,
  Dominican Republic. Association for Computational Linguistics.

\bibitem[{Katharopoulos et~al.(2020)Katharopoulos, Vyas, Pappas, and
  Fleuret}]{katharopoulos_et_al_2020}
A.~Katharopoulos, A.~Vyas, N.~Pappas, and F.~Fleuret. 2020.
\newblock Transformers are {RNNs}: Fast autoregressive transformers with linear
  attention.
\newblock In \emph{Proceedings of the International Conference on Machine
  Learning (ICML)}.

\bibitem[{Katiyar and Cardie(2018)}]{katiyar-cardie-2018-nested}
Arzoo Katiyar and Claire Cardie. 2018.
\newblock \href {https://doi.org/10.18653/v1/N18-1079} {Nested named entity
  recognition revisited}.
\newblock In \emph{Proceedings of the 2018 Conference of the North {A}merican
  Chapter of the Association for Computational Linguistics: Human Language
  Technologies, Volume 1 (Long Papers)}, pages 861--871, New Orleans,
  Louisiana. Association for Computational Linguistics.

\bibitem[{Khosla et~al.(2021)Khosla, Yu, Manuvinakurike, Ng, Poesio, Strube,
  and Ros{\'e}}]{khosla-etal-2021-codi}
Sopan Khosla, Juntao Yu, Ramesh Manuvinakurike, Vincent Ng, Massimo Poesio,
  Michael Strube, and Carolyn Ros{\'e}. 2021.
\newblock \href {https://doi.org/10.18653/v1/2021.codi-sharedtask.1} {The
  {CODI}-{CRAC} 2021 shared task on anaphora, bridging, and discourse deixis in
  dialogue}.
\newblock In \emph{Proceedings of the CODI-CRAC 2021 Shared Task on Anaphora,
  Bridging, and Discourse Deixis in Dialogue}, pages 1--15, Punta Cana,
  Dominican Republic. Association for Computational Linguistics.

\bibitem[{Kingma and Ba(2015)}]{KingmaB14}
Diederik~P. Kingma and Jimmy Ba. 2015.
\newblock \href {http://arxiv.org/abs/1412.6980} {Adam: {A} method for
  stochastic optimization}.
\newblock In \emph{3rd International Conference on Learning Representations,
  {ICLR} 2015, San Diego, CA, USA, May 7-9, 2015, Conference Track
  Proceedings}.

\bibitem[{Kirstain et~al.(2021)Kirstain, Ram, and
  Levy}]{kirstain-etal-2021-coreference}
Yuval Kirstain, Ori Ram, and Omer Levy. 2021.
\newblock \href {https://doi.org/10.18653/v1/2021.acl-short.3} {Coreference
  resolution without span representations}.
\newblock In \emph{Proceedings of the 59th Annual Meeting of the Association
  for Computational Linguistics and the 11th International Joint Conference on
  Natural Language Processing (Volume 2: Short Papers)}, pages 14--19, Online.
  Association for Computational Linguistics.

\bibitem[{Klenner and Tuggener(2011)}]{klenner-tuggener-2011-incremental}
Manfred Klenner and Don Tuggener. 2011.
\newblock \href {https://aclanthology.org/R11-1025} {An incremental
  entity-mention model for coreference resolution with restrictive antecedent
  accessibility}.
\newblock In \emph{Proceedings of the International Conference Recent Advances
  in Natural Language Processing 2011}, pages 178--185, Hissar, Bulgaria.
  Association for Computational Linguistics.

\bibitem[{Lee et~al.(2017)Lee, He, Lewis, and Zettlemoyer}]{lee-etal-2017-end}
Kenton Lee, Luheng He, Mike Lewis, and Luke Zettlemoyer. 2017.
\newblock \href {https://doi.org/10.18653/v1/D17-1018} {End-to-end neural
  coreference resolution}.
\newblock In \emph{Proceedings of the 2017 Conference on Empirical Methods in
  Natural Language Processing}, pages 188--197, Copenhagen, Denmark.
  Association for Computational Linguistics.

\bibitem[{Lee et~al.(2018)Lee, He, and Zettlemoyer}]{lee-etal-2018-higher}
Kenton Lee, Luheng He, and Luke Zettlemoyer. 2018.
\newblock \href {https://doi.org/10.18653/v1/N18-2108} {Higher-order
  coreference resolution with coarse-to-fine inference}.
\newblock In \emph{Proceedings of the 2018 Conference of the North {A}merican
  Chapter of the Association for Computational Linguistics: Human Language
  Technologies, Volume 2 (Short Papers)}, pages 687--692, New Orleans,
  Louisiana. Association for Computational Linguistics.

\bibitem[{Liu et~al.(2019)Liu, Zettlemoyer, and
  Eisenstein}]{liu-etal-2019-referential}
Fei Liu, Luke Zettlemoyer, and Jacob Eisenstein. 2019.
\newblock \href {https://doi.org/10.18653/v1/P19-1593} {The referential reader:
  A recurrent entity network for anaphora resolution}.
\newblock In \emph{Proceedings of the 57th Annual Meeting of the Association
  for Computational Linguistics}, pages 5918--5925, Florence, Italy.
  Association for Computational Linguistics.

\bibitem[{Liu et~al.(2022)Liu, Jiang, Monath, Cotterell, and
  Sachan}]{liu-etal-2022-transition}
Tianyu Liu, Yuchen~Eleanor Jiang, Nicholas Monath, Ryan Cotterell, and Mrinmaya
  Sachan. 2022.
\newblock Transition-based structured prediction with autoregressive models.
\newblock In \emph{Proceedings of the 2022 Conference on Empirical Methods in
  Natural Language Processing}, Abu Dhabi, United Arab Emirates and Virtual.
  Association for Computational Lingustics.

\bibitem[{Logan~IV et~al.(2021)Logan~IV, McCallum, Singh, and
  Bikel}]{logan-iv-etal-2021-benchmarking}
Robert~L Logan~IV, Andrew McCallum, Sameer Singh, and Dan Bikel. 2021.
\newblock \href {https://doi.org/10.18653/v1/2021.acl-long.364} {Benchmarking
  scalable methods for streaming cross document entity coreference}.
\newblock In \emph{Proceedings of the 59th Annual Meeting of the Association
  for Computational Linguistics and the 11th International Joint Conference on
  Natural Language Processing (Volume 1: Long Papers)}, pages 4717--4731,
  Online. Association for Computational Linguistics.

\bibitem[{Loshchilov and Hutter(2019)}]{loshchilov2018decoupled}
Ilya Loshchilov and Frank Hutter. 2019.
\newblock \href {https://openreview.net/forum?id=Bkg6RiCqY7} {Decoupled weight
  decay regularization}.
\newblock In \emph{International Conference on Learning Representations}.

\bibitem[{Luo(2005)}]{luo-2005-coreference}
Xiaoqiang Luo. 2005.
\newblock \href {https://aclanthology.org/H05-1004} {On coreference resolution
  performance metrics}.
\newblock In \emph{Proceedings of Human Language Technology Conference and
  Conference on Empirical Methods in Natural Language Processing}, pages
  25--32, Vancouver, British Columbia, Canada. Association for Computational
  Linguistics.

\bibitem[{Madureira and Schlangen(2020)}]{madureira-schlangen-2020-incremental}
Brielen Madureira and David Schlangen. 2020.
\newblock \href {https://doi.org/10.18653/v1/2020.emnlp-main.26} {Incremental
  processing in the age of non-incremental encoders: An empirical assessment of
  bidirectional models for incremental {NLU}}.
\newblock In \emph{Proceedings of the 2020 Conference on Empirical Methods in
  Natural Language Processing (EMNLP)}, pages 357--374, Online. Association for
  Computational Linguistics.

\bibitem[{Martin et~al.(2020)Martin, Poddar, and
  Upasani}]{martin-etal-2020-mudoco}
Scott Martin, Shivani Poddar, and Kartikeya Upasani. 2020.
\newblock \href {https://aclanthology.org/2020.lrec-1.13} {{M}u{D}o{C}o: Corpus
  for multidomain coreference resolution and referring expression generation}.
\newblock In \emph{Proceedings of the 12th Language Resources and Evaluation
  Conference}, pages 104--111, Marseille, France. European Language Resources
  Association.

\bibitem[{Paszke et~al.(2019)Paszke, Gross, Massa, Lerer, Bradbury, Chanan,
  Killeen, Lin, Gimelshein, Antiga, Desmaison, Kopf, Yang, DeVito, Raison,
  Tejani, Chilamkurthy, Steiner, Fang, Bai, and Chintala}]{pytorch}
Adam Paszke, Sam Gross, Francisco Massa, Adam Lerer, James Bradbury, Gregory
  Chanan, Trevor Killeen, Zeming Lin, Natalia Gimelshein, Luca Antiga, Alban
  Desmaison, Andreas Kopf, Edward Yang, Zachary DeVito, Martin Raison, Alykhan
  Tejani, Sasank Chilamkurthy, Benoit Steiner, Lu~Fang, Junjie Bai, and Soumith
  Chintala. 2019.
\newblock \href
  {http://papers.neurips.cc/paper/9015-pytorch-an-imperative-style-high-performance-deep-learning-library.pdf}
  {Pytorch: An imperative style, high-performance deep learning library}.
\newblock In H.~Wallach, H.~Larochelle, A.~Beygelzimer, F.~d\textquotesingle
  Alch\'{e}-Buc, E.~Fox, and R.~Garnett, editors, \emph{Advances in Neural
  Information Processing Systems 32}, pages 8024--8035. Curran Associates, Inc.

\bibitem[{Pradhan et~al.(2012)Pradhan, Moschitti, Xue, Uryupina, and
  Zhang}]{pradhan-etal-2012-conll}
Sameer Pradhan, Alessandro Moschitti, Nianwen Xue, Olga Uryupina, and Yuchen
  Zhang. 2012.
\newblock \href {https://aclanthology.org/W12-4501} {{C}o{NLL}-2012 shared
  task: Modeling multilingual unrestricted coreference in {O}nto{N}otes}.
\newblock In \emph{Joint Conference on {EMNLP} and {C}o{NLL} - Shared Task},
  pages 1--40, Jeju Island, Korea. Association for Computational Linguistics.

\bibitem[{Rudinger et~al.(2018)Rudinger, Naradowsky, Leonard, and
  Van~Durme}]{rudinger-etal-2018-gender}
Rachel Rudinger, Jason Naradowsky, Brian Leonard, and Benjamin Van~Durme. 2018.
\newblock \href {https://doi.org/10.18653/v1/N18-2002} {Gender bias in
  coreference resolution}.
\newblock In \emph{Proceedings of the 2018 Conference of the North {A}merican
  Chapter of the Association for Computational Linguistics: Human Language
  Technologies, Volume 2 (Short Papers)}, pages 8--14, New Orleans, Louisiana.
  Association for Computational Linguistics.

\bibitem[{Subramanian and Roth(2019)}]{subramanian-roth-2019-improving}
Sanjay Subramanian and Dan Roth. 2019.
\newblock \href {https://doi.org/10.18653/v1/S19-1021} {Improving
  generalization in coreference resolution via adversarial training}.
\newblock In \emph{Proceedings of the Eighth Joint Conference on Lexical and
  Computational Semantics (*{SEM} 2019)}, pages 192--197, Minneapolis,
  Minnesota. Association for Computational Linguistics.

\bibitem[{Toshniwal et~al.(2020{\natexlab{a}})Toshniwal, Ettinger, Gimpel, and
  Livescu}]{toshniwal-etal-2020-petra}
Shubham Toshniwal, Allyson Ettinger, Kevin Gimpel, and Karen Livescu.
  2020{\natexlab{a}}.
\newblock \href {https://doi.org/10.18653/v1/2020.acl-main.481} {{P}e{T}ra: {A}
  {S}parsely {S}upervised {M}emory {M}odel for {P}eople {T}racking}.
\newblock In \emph{Proceedings of the 58th Annual Meeting of the Association
  for Computational Linguistics}, pages 5415--5428, Online. Association for
  Computational Linguistics.

\bibitem[{Toshniwal et~al.(2020{\natexlab{b}})Toshniwal, Wiseman, Ettinger,
  Livescu, and Gimpel}]{toshniwal-etal-2020-learning}
Shubham Toshniwal, Sam Wiseman, Allyson Ettinger, Karen Livescu, and Kevin
  Gimpel. 2020{\natexlab{b}}.
\newblock \href {https://doi.org/10.18653/v1/2020.emnlp-main.685} {Learning to
  {I}gnore: {L}ong {D}ocument {C}oreference with {B}ounded {M}emory {N}eural
  {N}etworks}.
\newblock In \emph{Proceedings of the 2020 Conference on Empirical Methods in
  Natural Language Processing (EMNLP)}, pages 8519--8526, Online. Association
  for Computational Linguistics.

\bibitem[{Toshniwal et~al.(2021)Toshniwal, Xia, Wiseman, Livescu, and
  Gimpel}]{toshniwal-etal-2021-generalization}
Shubham Toshniwal, Patrick Xia, Sam Wiseman, Karen Livescu, and Kevin Gimpel.
  2021.
\newblock \href {https://doi.org/10.18653/v1/2021.crac-1.12} {On generalization
  in coreference resolution}.
\newblock In \emph{Proceedings of the Fourth Workshop on Computational Models
  of Reference, Anaphora and Coreference}, pages 111--120, Punta Cana,
  Dominican Republic. Association for Computational Linguistics.

\bibitem[{Urbanek et~al.(2019)Urbanek, Fan, Karamcheti, Jain, Humeau, Dinan,
  Rockt{\"a}schel, Kiela, Szlam, and Weston}]{urbanek-etal-2019-learning}
Jack Urbanek, Angela Fan, Siddharth Karamcheti, Saachi Jain, Samuel Humeau,
  Emily Dinan, Tim Rockt{\"a}schel, Douwe Kiela, Arthur Szlam, and Jason
  Weston. 2019.
\newblock \href {https://doi.org/10.18653/v1/D19-1062} {Learning to speak and
  act in a fantasy text adventure game}.
\newblock In \emph{Proceedings of the 2019 Conference on Empirical Methods in
  Natural Language Processing and the 9th International Joint Conference on
  Natural Language Processing (EMNLP-IJCNLP)}, pages 673--683, Hong Kong,
  China. Association for Computational Linguistics.

\bibitem[{Vilain et~al.(1995)Vilain, Burger, Aberdeen, Connolly, and
  Hirschman}]{vilain-etal-1995-model}
Marc Vilain, John Burger, John Aberdeen, Dennis Connolly, and Lynette
  Hirschman. 1995.
\newblock \href {https://aclanthology.org/M95-1005} {A model-theoretic
  coreference scoring scheme}.
\newblock In \emph{Sixth Message Understanding Conference ({MUC}-6):
  Proceedings of a Conference Held in {C}olumbia, {M}aryland, November 6-8,
  1995}.

\bibitem[{Wang et~al.(2018)Wang, Lu, Wang, and
  Jin}]{wang-etal-2018-neural-transition}
Bailin Wang, Wei Lu, Yu~Wang, and Hongxia Jin. 2018.
\newblock \href {https://doi.org/10.18653/v1/D18-1124} {A neural
  transition-based model for nested mention recognition}.
\newblock In \emph{Proceedings of the 2018 Conference on Empirical Methods in
  Natural Language Processing}, pages 1011--1017, Brussels, Belgium.
  Association for Computational Linguistics.

\bibitem[{Wang and Komatsuzaki(2021)}]{gpt-j}
Ben Wang and Aran Komatsuzaki. 2021.
\newblock \href {https://github.com/kingoflolz/mesh-transformer-jax}
  {{GPT-J-6B: A 6 Billion Parameter Autoregressive Language Model}}.
\newblock GitHub.

\bibitem[{Wang et~al.(2019)Wang, Shi, Kim, Oh, Yang, Zhang, and
  Yu}]{wang-etal-2019-persuasion}
Xuewei Wang, Weiyan Shi, Richard Kim, Yoojung Oh, Sijia Yang, Jingwen Zhang,
  and Zhou Yu. 2019.
\newblock \href {https://doi.org/10.18653/v1/P19-1566} {Persuasion for good:
  Towards a personalized persuasive dialogue system for social good}.
\newblock In \emph{Proceedings of the 57th Annual Meeting of the Association
  for Computational Linguistics}, pages 5635--5649, Florence, Italy.
  Association for Computational Linguistics.

\bibitem[{Webster and Curran(2014)}]{webster-curran-2014-limited}
Kellie Webster and James~R. Curran. 2014.
\newblock \href {https://aclanthology.org/C14-1201} {Limited memory incremental
  coreference resolution}.
\newblock In \emph{Proceedings of {COLING} 2014, the 25th International
  Conference on Computational Linguistics: Technical Papers}, pages 2129--2139,
  Dublin, Ireland. Dublin City University and Association for Computational
  Linguistics.

\bibitem[{Webster et~al.(2018)Webster, Recasens, Axelrod, and
  Baldridge}]{webster-etal-2018-mind}
Kellie Webster, Marta Recasens, Vera Axelrod, and Jason Baldridge. 2018.
\newblock \href {https://doi.org/10.1162/tacl_a_00240} {Mind the {GAP}: A
  balanced corpus of gendered ambiguous pronouns}.
\newblock \emph{Transactions of the Association for Computational Linguistics},
  6:605--617.

\bibitem[{Weischedel et~al.(2013)Weischedel, Palmer, Marcus, Hovy, Pradhan,
  Ramshaw, Xue, Taylor, Kaufman, Franchini, El-Bachouti, Belvin, and
  Houston}]{ontonotes}
Ralph Weischedel, Martha Palmer, Mitchell Marcus, Eduard Hovy, Sameer Pradhan,
  Lance Ramshaw, Nianwen Xue, Ann Taylor, Jeff Kaufman, Michelle Franchini,
  Mohammed El-Bachouti, Robert Belvin, and Ann Houston. 2013.
\newblock Ontonotes release 5.0.
\newblock \emph{LDC2013T19, Philadelphia, Penn.: Linguistic Data Consortium}.

\bibitem[{Wu et~al.(2020)Wu, Wang, Yuan, Wu, and Li}]{wu-etal-2020-corefqa}
Wei Wu, Fei Wang, Arianna Yuan, Fei Wu, and Jiwei Li. 2020.
\newblock \href {https://doi.org/10.18653/v1/2020.acl-main.622} {{C}oref{QA}:
  Coreference resolution as query-based span prediction}.
\newblock In \emph{Proceedings of the 58th Annual Meeting of the Association
  for Computational Linguistics}, pages 6953--6963, Online. Association for
  Computational Linguistics.

\bibitem[{Xia et~al.(2020)Xia, Sedoc, and
  Van~Durme}]{xia-etal-2020-incremental}
Patrick Xia, Jo{\~a}o Sedoc, and Benjamin Van~Durme. 2020.
\newblock \href {https://doi.org/10.18653/v1/2020.emnlp-main.695} {Incremental
  neural coreference resolution in constant memory}.
\newblock In \emph{Proceedings of the 2020 Conference on Empirical Methods in
  Natural Language Processing (EMNLP)}, pages 8617--8624, Online. Association
  for Computational Linguistics.

\bibitem[{Xia and Van~Durme(2021)}]{xia-van-durme-2021-moving}
Patrick Xia and Benjamin Van~Durme. 2021.
\newblock \href {https://doi.org/10.18653/v1/2021.emnlp-main.425} {Moving on
  from {O}nto{N}otes: Coreference resolution model transfer}.
\newblock In \emph{Proceedings of the 2021 Conference on Empirical Methods in
  Natural Language Processing}, pages 5241--5256, Online and Punta Cana,
  Dominican Republic. Association for Computational Linguistics.

\bibitem[{Xu and Choi(2020)}]{xu-choi-2020-revealing}
Liyan Xu and Jinho~D. Choi. 2020.
\newblock \href {https://doi.org/10.18653/v1/2020.emnlp-main.686} {Revealing
  the myth of higher-order inference in coreference resolution}.
\newblock In \emph{Proceedings of the 2020 Conference on Empirical Methods in
  Natural Language Processing (EMNLP)}, pages 8527--8533, Online. Association
  for Computational Linguistics.

\bibitem[{Yang et~al.(2019)Yang, Dai, Yang, Carbonell, Salakhutdinov, and
  Le}]{xlnet}
Zhilin Yang, Zihang Dai, Yiming Yang, Jaime Carbonell, Russ~R Salakhutdinov,
  and Quoc~V Le. 2019.
\newblock \href
  {https://proceedings.neurips.cc/paper/2019/file/dc6a7e655d7e5840e66733e9ee67cc69-Paper.pdf}
  {Xlnet: Generalized autoregressive pretraining for language understanding}.
\newblock In \emph{Advances in Neural Information Processing Systems},
  volume~32. Curran Associates, Inc.

\bibitem[{Yu et~al.(2020{\natexlab{a}})Yu, Bohnet, and
  Poesio}]{yu-etal-2020-named}
Juntao Yu, Bernd Bohnet, and Massimo Poesio. 2020{\natexlab{a}}.
\newblock \href {https://doi.org/10.18653/v1/2020.acl-main.577} {Named entity
  recognition as dependency parsing}.
\newblock In \emph{Proceedings of the 58th Annual Meeting of the Association
  for Computational Linguistics}, pages 6470--6476, Online. Association for
  Computational Linguistics.

\bibitem[{Yu et~al.(2022)Yu, Khosla, Moosavi, Paun, Pradhan, and
  Poesio}]{yu-etal-2022-universal}
Juntao Yu, Sopan Khosla, Nafise Moosavi, Silviu Paun, Sameer Pradhan, and
  Massimo Poesio. 2022.
\newblock The universal anaphora scorer.
\newblock In \emph{Proceedings of the 13th Language Resources and Evaluation
  Conference}, Marseille, France. European Language Resources Association.

\bibitem[{Yu et~al.(2020{\natexlab{b}})Yu, Uma, and
  Poesio}]{yu-etal-2020-cluster}
Juntao Yu, Alexandra Uma, and Massimo Poesio. 2020{\natexlab{b}}.
\newblock \href {https://aclanthology.org/2020.lrec-1.2} {A cluster ranking
  model for full anaphora resolution}.
\newblock In \emph{Proceedings of the 12th Language Resources and Evaluation
  Conference}, pages 11--20, Marseille, France. European Language Resources
  Association.

\bibitem[{Yuan et~al.(2022)Yuan, Xia, May, Van~Durme, and
  Boyd-Graber}]{yuan-etal-2022-adapting}
Michelle Yuan, Patrick Xia, Chandler May, Benjamin Van~Durme, and Jordan
  Boyd-Graber. 2022.
\newblock \href {https://doi.org/10.18653/v1/2022.acl-long.519} {Adapting
  coreference resolution models through active learning}.
\newblock In \emph{Proceedings of the 60th Annual Meeting of the Association
  for Computational Linguistics (Volume 1: Long Papers)}, pages 7533--7549,
  Dublin, Ireland. Association for Computational Linguistics.

\bibitem[{Zhao et~al.(2018)Zhao, Wang, Yatskar, Ordonez, and
  Chang}]{zhao-etal-2018-gender}
Jieyu Zhao, Tianlu Wang, Mark Yatskar, Vicente Ordonez, and Kai-Wei Chang.
  2018.
\newblock \href {https://doi.org/10.18653/v1/N18-2003} {Gender bias in
  coreference resolution: Evaluation and debiasing methods}.
\newblock In \emph{Proceedings of the 2018 Conference of the North {A}merican
  Chapter of the Association for Computational Linguistics: Human Language
  Technologies, Volume 2 (Short Papers)}, pages 15--20, New Orleans, Louisiana.
  Association for Computational Linguistics.

\end{thebibliography}
\bibliographystyle{acl_natbib}

\appendix

\section{Action Constraints}
To ensure the final state is always reached, it is necessary to enforce a set of rules during mention detection: 
\begin{enumerate}
    \item \adv\ can only be called on the final token if the stack is empty. 
    \item \pop\ and \peek\ can only be called if the stack is non-empty.
    \item \push\ can only be called once per token, ensuring that left boundaries are only marked once.
    \item \push\ cannot directly follow \pop\ or \peek. Allowing this action sequence would either admit multiple paths to the same mention or non-nested overlapping mentions.
    \item \pop\ cannot directly follow \peek, or else the same mention would be proposed twice.
    \item \peek\ cannot be called on the final token. This action would imply the stack is non-empty on the final token, and that \pop\ must be called. 
\end{enumerate}

\section{$k$-Sentence-Incremental Mention Detection}

We repeat the experiment in Section \ref{sec:analysis} for mention detection.
For each $k$-Sentence-Incremental setting, we evaluate on the dev set and record the mention detection recall, precision and F1.

\begin{figure}[ht]
    \centering
    \resizebox{\columnwidth}{!}{
    \begin{tikzpicture}
\definecolor{darkslategray38}{RGB}{38,38,38}
\definecolor{lavender234234242}{RGB}{234,234,242}
\definecolor{lightgray204}{RGB}{204,204,204}
\definecolor{mediumseagreen85168104}{RGB}{85,168,104}
\definecolor{peru22113282}{RGB}{221,132,82}
\definecolor{steelblue76114176}{RGB}{76,114,176}

\begin{axis}[
axis background/.style={fill=lavender234234242},
axis line style={white},
legend cell align={left},
legend style={
  fill opacity=0.8,
  draw opacity=1,
  text opacity=1,
  at={(0.97,0.03)},
  anchor=south east,
  draw=lightgray204,
  fill=lavender234234242
},
legend entries={Recall, Precision, F1},
tick align=outside,
x grid style={white},
xlabel=\textcolor{darkslategray38}{\(\displaystyle k\)-Sentence Setting},
xmajorgrids,
xmajorticks=true,
xtick pos=bottom,
xmin=-0.15, xmax=25.15,
xtick style={color=darkslategray38},
y grid style={white},
ylabel=\textcolor{darkslategray38}{Score},
ymajorgrids,
ymajorticks=true,
ymin=81.6595, ymax=87.6105,
ytick style={color=darkslategray38},
ytick pos=left
]
\addplot [very thick, steelblue76114176]
table[row sep=crcr] {%
1 81.93 \\
4 82.22 \\
8 83.67 \\
12 84.75 \\
16 84.48 \\
20 85.27 \\
24 85.14 \\
};
\addplot [very thick, peru22113282]
table[row sep=crcr] {%
1 82.17 \\
4 85.38 \\
8 85.94 \\
12 85.79 \\
16 87.34 \\
20 86.83 \\
24 87.15 \\
};
\addplot [very thick, mediumseagreen85168104]
table[row sep=crcr] {%
1 82.05 \\
4 83.77 \\
8 84.79 \\
12 85.27 \\
16 85.89 \\
20 86.04 \\
24 86.14 \\
};
\end{axis}

\end{tikzpicture}
    }
    \caption{Mention Detection performance (average of \muc, \bcub\ and \ceaf) of each $k$-Sentence-Incremental model on the OntoNotes dev set.}
    \label{fig:kscores_md}
\end{figure}
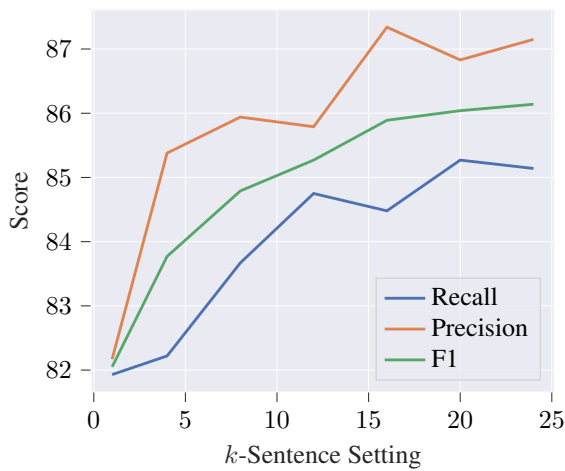

The results are shown in Figure \ref{fig:kscores_md}.
Certain trends remain the same as for the CoNLL score, namely that performance rises more when $k$ is small compared to when it is large. 
However, we do not see the same dramatic difference in recall between $k=1$ to $k=24$ settings as in Section \ref{sec:analysis}.
Here, the difference in recall between the two settings is around 3\%, whereas in Figure \ref{fig:kscores} it is 9.7\%.

Overall, the reduced severity between $k=1$ and $k=24$ settings compared to Figure \ref{fig:kscores} most likely indicate that XLNet's caching deficiencies affect mention clustering (particularly false negatives) more seriously than mention detection.

\section{Partitioning Document Clusters}

\begin{figure*}[t]
    \centering
    \begin{subfigure}[t]{\columnwidth}
    \resizebox{\columnwidth}{!}{
    \begin{tikzpicture}
    
    \definecolor{darkslategray38}{RGB}{38,38,38};
    \definecolor{gray14712096}{RGB}{147,120,96};
    \definecolor{indianred1967882}{RGB}{196,78,82};
    \definecolor{lavender234234242}{RGB}{234,234,242};
    \definecolor{lightgray204}{RGB}{204,204,204};
    \definecolor{lightslategray129114179}{RGB}{129,114,179};
    \definecolor{mediumseagreen85168104}{RGB}{85,168,104};
    \definecolor{orchid218139195}{RGB}{218,139,195};
    \definecolor{peru22113282}{RGB}{221,132,82};
    \definecolor{steelblue76114176}{RGB}{76,114,176};
    
    \begin{axis}[
    axis background/.style={fill=lavender234234242},
    axis line style={white},
    legend columns=4,
    legend style={
      fill opacity=0.8,
      anchor=south,
      at={(0.5,1.03)},
      draw opacity=1,
      text opacity=1,
      draw=lightgray204,
      fill=lavender234234242,
      font=\small
    },
    legend entries={$k=1$,$k=4$,$k=8$,$k=12$,$k=16$,$k=20$,$k=24$},
    tick align=outside,
    x grid style={white},
    xlabel=\textcolor{darkslategray38}{Partition Size},
    xmajorgrids,
    xmajorticks=true,
    xmin=-1.35, xmax=50.35,
    xtick style={color=darkslategray38},
    xtick pos={bottom},
    y grid style={white},
    ylabel=\textcolor{darkslategray38}{Recall},
    ymajorgrids,
    ymajorticks=true,
    ymin=66.540256892757, ymax=88.4333682728814,
    ytick style={color=darkslategray38},
    ytick pos={left}
    ]
    \addplot [very thick, steelblue76114176]
    table[row sep=crcr] {%
    1 86.2381380564032 \\
    2 79.7438694828602 \\
    3 77.7204599801313 \\
    4 75.8470765260284 \\
    8 72.8873953744037 \\
    12 71.5158599547277 \\
    16 70.3506045699479 \\
    20 69.8135253767202 \\
    24 69.243138892126 \\
    28 68.7787770483136 \\
    32 68.3822395876299 \\
    36 67.9901231399478 \\
    40 67.9203822043913 \\
    44 67.7556706127535 \\
    48 67.5353983191263 \\
    };
    \addplot [very thick, peru22113282]
    table[row sep=crcr] {%
    1 86.9219460903742 \\
    2 85.2091609317874 \\
    3 82.5167113330094 \\
    4 83.3004721138449 \\
    8 79.4108115505048 \\
    12 77.719539786021 \\
    16 76.4190443790405 \\
    20 75.7639064071199 \\
    24 75.1006564193584 \\
    28 74.5330656157426 \\
    32 74.031944447216 \\
    36 73.827320277907 \\
    40 73.7252364009913 \\
    44 73.6181104646492 \\
    48 73.4120340332108 \\
    };
    \addplot [very thick, mediumseagreen85168104]
    table[row sep=crcr] {%
    1 86.9325891086488 \\
    2 85.1709072808626 \\
    3 83.287836069048 \\
    4 83.3398115922159 \\
    8 81.6122612135934 \\
    12 79.1873111434029 \\
    16 78.2919227446731 \\
    20 77.152406274684 \\
    24 76.8443700904663 \\
    28 76.0913860872231 \\
    32 75.6678511765502 \\
    36 75.3026164764875 \\
    40 75.2855405840067 \\
    44 75.0896265522117 \\
    48 74.960496262111 \\
    };
    \addplot [very thick, indianred1967882]
    table[row sep=crcr] {%
    1 86.9418689228373 \\
    2 84.9507439286127 \\
    3 83.9781952663234 \\
    4 83.0556586755554 \\
    8 80.8282996343307 \\
    12 80.8098504689344 \\
    16 78.3742455588308 \\
    20 77.6920367298017 \\
    24 77.3978194476081 \\
    28 76.4974275813329 \\
    32 76.000107398333 \\
    36 76.0112676336055 \\
    40 75.6573497281233 \\
    44 75.5021005215566 \\
    48 75.4386318838711 \\
    };
    \addplot [very thick, lightslategray129114179]
    table[row sep=crcr] {%
    1 87.4382268465121 \\
    2 85.8172930059633 \\
    3 84.6124305150217 \\
    4 83.9919865104355 \\
    8 82.4642771661078 \\
    12 80.9638154502684 \\
    16 81.1579170022215 \\
    20 79.4684006401296 \\
    24 78.913873556388 \\
    28 78.2796258245785 \\
    32 78.0987503907159 \\
    36 77.5020249595352 \\
    40 77.4004185860774 \\
    44 77.2760895046652 \\
    48 77.2493334906589 \\
    };
    \addplot [very thick, gray14712096]
    table[row sep=crcr] {%
    1 86.5667791276086 \\
    2 85.4170318916889 \\
    3 84.0188969508511 \\
    4 83.8163171461018 \\
    8 81.7247171432261 \\
    12 80.7408433126407 \\
    16 80.2981502966503 \\
    20 80.605200022547 \\
    24 79.0034032325806 \\
    28 78.5110301248781 \\
    32 78.1282142859596 \\
    36 77.7085110964955 \\
    40 77.8337415261951 \\
    44 77.5331041723338 \\
    48 77.2907264880714 \\
    };
    \addplot [very thick, orchid218139195]
    table[row sep=crcr] {%
    1 87.0951221126295 \\
    2 85.3117567132715 \\
    3 84.410825313377 \\
    4 83.6511082004442 \\
    8 82.2190867340505 \\
    12 81.3598746150779 \\
    16 80.2581488627037 \\
    20 79.854712649481 \\
    24 80.0435794826503 \\
    28 78.887013209873 \\
    32 78.3212012819135 \\
    36 77.9013412354894 \\
    40 77.8517606801543 \\
    44 77.7153082553119 \\
    48 77.7307876436267 \\
    };
    \end{axis}
    \end{tikzpicture}
    }
    \caption{}
    \label{fig:cpartition}
    \end{subfigure}
    \begin{subfigure}[t]{\columnwidth}
    \resizebox{\columnwidth}{!}{
    
\begin{tikzpicture}

\definecolor{darkslategray38}{RGB}{38,38,38}
\definecolor{gray14712096}{RGB}{147,120,96}
\definecolor{indianred1967882}{RGB}{196,78,82}
\definecolor{lavender234234242}{RGB}{234,234,242}
\definecolor{lightgray204}{RGB}{204,204,204}
\definecolor{lightslategray129114179}{RGB}{129,114,179}
\definecolor{mediumseagreen85168104}{RGB}{85,168,104}
\definecolor{orchid218139195}{RGB}{218,139,195}
\definecolor{peru22113282}{RGB}{221,132,82}
\definecolor{steelblue76114176}{RGB}{76,114,176}

\begin{axis}[
axis background/.style={fill=lavender234234242},
axis line style={white},
legend cell align={left},
legend columns=4,
legend style={
  fill opacity=0.8,
  draw opacity=1,
  text opacity=1,
  anchor=south,
  at={(0.5,1.03)},
  draw=lightgray204,
  fill=lavender234234242,
  font=\small
},
legend entries={$k=1$,$k=4$,$k=8$,$k=12$,$k=16$,$k=20$,$k=24$},
tick align=outside,
x grid style={white},
xlabel=\textcolor{darkslategray38}{Partition Size},
xmajorgrids,
xmajorticks=true,
xmin=-1.35, xmax=50.35,
xtick style={color=darkslategray38},
xtick pos={bottom},
y grid style={white},
ylabel=\textcolor{darkslategray38}{Recall},
ymajorgrids,
ymajorticks=true,
ymin=79.5399818420951, ymax=90.2451766376397,
ytick style={color=darkslategray38},
ytick pos={left}
]
\addplot [very thick, steelblue76114176]
table[row sep=crcr]{%
1 88.5260482846252 \\
2 83.2186956161432 \\
3 82.3081938708976 \\
4 81.3686682321369 \\
8 80.8336323406291 \\
12 80.6248928999829 \\
16 80.4550270451124 \\
20 80.3485313459009 \\
24 80.2832953435363 \\
28 80.1798890504659 \\
32 80.1400994599219 \\
36 80.026581605529 \\
40 80.1335736245097 \\
44 80.1746493781424 \\
48 80.1109057301294 \\
};
\addplot [very thick, peru22113282]
table[row sep=crcr]{%
1 89.2503176620076 \\
2 88.3069528466064 \\
3 86.6188509744259 \\
4 87.9369339147937 \\
8 86.0064585575888 \\
12 85.3829896612783 \\
16 84.8993475715162 \\
20 84.6065322227093 \\
24 84.5110170411375 \\
28 84.2570151343782 \\
32 84.0757178760494 \\
36 84.0191387559809 \\
40 84.0771758719389 \\
44 84.112198994443 \\
48 84.0137311856351 \\
};
\addplot [very thick, mediumseagreen85168104]
table[row sep=crcr]{%
1 89.1232528589581 \\
2 88.2738045910334 \\
3 87.3143519524741 \\
4 88.0040254948004 \\
8 87.8363832077503 \\
12 86.4168618266979 \\
16 86.2934255283556 \\
20 85.6587023235423 \\
24 85.6995549766634 \\
28 85.3449668767168 \\
32 85.1719159403241 \\
36 84.9760765550239 \\
40 85.1107812996926 \\
44 85.0965863985181 \\
48 85.0171639820438 \\
};
\addplot [very thick, indianred1967882]
table[row sep=crcr]{%
1 89.0978398983482 \\
2 87.9754702908759 \\
3 87.7780192711729 \\
4 87.5410935927541 \\
8 86.9872024877407 \\
12 87.5307020049123 \\
16 86.1317124853622 \\
20 85.7189829022359 \\
24 85.8135243677412 \\
28 85.3934399741477 \\
32 85.1612213250628 \\
36 85.2631578947368 \\
40 85.0630764337963 \\
44 85.0436623445356 \\
48 85.0118827567996 \\
};
\addplot [very thick, lightslategray129114179]
table[row sep=crcr]{%
1 89.7585768742058 \\
2 89.019640341427 \\
3 88.4662754473665 \\
4 88.5072123448507 \\
8 88.5240999880397 \\
12 87.8677100588336 \\
16 88.5964423130541 \\
20 87.4506795265235 \\
24 87.3602518180831 \\
28 87.0469111865137 \\
32 87.0755574568205 \\
36 86.7676767676768 \\
40 86.7592494434432 \\
44 86.7372320719767 \\
48 86.7863744388698 \\
};
\addplot [very thick, gray14712096]
table[row sep=crcr]{%
1 88.6912325285896 \\
2 88.3981105494323 \\
3 87.7780192711729 \\
4 88.1516269708152 \\
8 87.6330582466212 \\
12 87.3707659793226 \\
16 87.5871298723025 \\
20 88.053485313459 \\
24 87.0563334418756 \\
28 86.8584046965046 \\
32 86.6852039997861 \\
36 86.4912280701754 \\
40 86.6744407929609 \\
44 86.5361206668431 \\
48 86.3850013203063 \\
};
\addplot [very thick, orchid218139195]
table[row sep=crcr]{%
1 89.2376111817027 \\
2 88.3069528466064 \\
3 88.1982177787438 \\
4 88.1382086548138 \\
8 88.2370529840928 \\
12 88.0219340835094 \\
16 87.6930797970222 \\
20 87.532880315651 \\
24 88.0332139368284 \\
28 87.2946625733829 \\
32 87.0060424576226 \\
36 86.8048910154173 \\
40 86.8440580939256 \\
44 86.8589573961365 \\
48 86.9184050699762 \\
};
\end{axis}

\end{tikzpicture}
}
    \caption{}
    \label{fig:mpartition}
    \end{subfigure}
    \caption{Evaluation results on the OntoNotes dev set when the gold labels and $k$-Sentence-Incremental predictions are partitioned according to various sizes. 
    Figure \ref{fig:cpartition} shows the CoNLL recall scores for coreference resolution. 
    Figure \ref{fig:mpartition} shows the mention detection recall scores.
    Notice that whenever $k$ is equal to the partition size, there is a noticeable performance increase, indicating that XLNet relies heavily on active inputs rather than its memory.}
    \label{fig:partition}
\end{figure*}
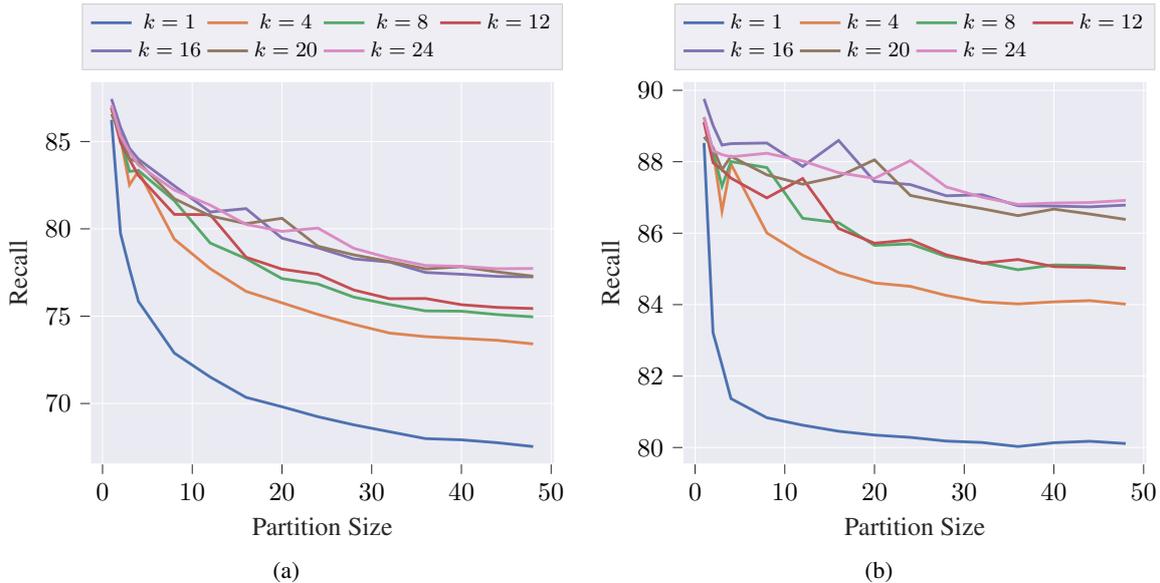

We explore the deficiency in the previous section further, guided by the hypothesis that XLNet relies on active inputs and cannot effectively use its memory.
We partition each document into segments of $k$ sentences, and call the number of sentences in each segment the `partition size'.
Within each segment, we maintain the original coreference links. 
However, we remove the coreference links between segments.
For each $k$-Sentence-Incremental model, we similarly partition their coreference predictions, and evaluate against the partitioned gold labels on the OntoNotes dev set.

Each segment therefore is independent from the other, and we can measure how reliant the model is on its active inputs by observing performance change across partition sizes.
For example, when the partition size is 1, the only coreference links are intra-sentential ones.
In this case, models are only evaluated on their intra-sentential coreference resolution ability.
When the partition size is large, the models are evaluated on the documents' original coreference chains.
Since the previous experiments demonstrated shifts in incrementality heavily affected recall, we measure the mention detection recall and CoNLL recall score.

The results are shown in Figure \ref{fig:partition}.
All models do well when the partition size is one, reflecting the fact that intra-sentential coreference is generally simpler than distantly linked mentions. 
As the partition size increases, model performance decreases as the coreference chains become more spread apart and raise the task difficulty.
Crucially, we notice a upward performance bump whenever the partition size matches the $k$-Sentence-Incremental setting, for both coreference performance and mention detection
When $k$ matches the partition size, the model observes coreference chains that are always within the active input window.
This performance bump therefore indicates XLNet is much better at mention detection and coreference when the coreference chain occurs within its active inputs.
Performance suffers whenever the model must rely more on its memory (whenever $k$ is not equal to the partition size).
In particular, these results suggest that more powerful pre-trained language models, in particular ones that can take better advantage of cached representations, may be more successful at incremental coreference resolution.

\begin{table*}[ht]
\small
\centering
\begin{tabular}{l|c c c c c c c c c|c}
\toprule																							
	&		\multicolumn{3}{c}{\muc}					&	\multicolumn{3}{c}{\bcub}					&	\multicolumn{3}{c}{\ceaf}					&	Avg.	\\
Model	&	Rec.	&	Prec.	&	F1	&	Rec.	&	Prec.	&	F1	&	Rec.	&	Prec.	&	F1	&	F1	\\
\midrule																					
Sent-Inc  & 76.4	&	78.7	&	77.6	&	68.0	&	68.2	&	68.1	&	56.1	&	70.6	&	62.6	&	69.4	\\
\indent\hspace{0.2cm}\textminus speaker & 76.4	&	79.0	&	77.7	&	68.0	&	68.6	&	68.3	&	56.3	&	70.8	&	62.7	&	69.5	\\
\bottomrule								
\end{tabular}
\caption{Results on the OntoNotes dev set comparing the Sent-Inc model with and without speaker embeddings.}
\label{tab:ablation_results}
\end{table*}

\begin{table}[ht]
    \small
    \centering
    \begin{tabular}{l|c c c c|c}
\toprule											
Model	&	\textsc{Light}	&	\textsc{AMI}	&	\textsc{Pers.}	&	\textsc{Swbd.}	&	Avg.	\\
\midrule
Sent-Inc	&	50.6	&	32.5	&	48.5	&	44.4	&	44.0	\\
\indent\hspace{0.1cm}\textminus speaker &	50.1	&	32.0	&	49.6	&	43.3	&	43.8	\\
\bottomrule			
    \end{tabular}
    \caption{Results on the CODI-CRAC dev set comparing the Sent-Inc model with and without speaker embeddings.}
    \label{tab:codi_ablation}
\end{table}

\section{Speaker Embeddings}
The ICoref-\textit{inc} model from \citet{xia-etal-2020-incremental} is an important comparison point as the only baseline in the sentence-incremental setting.
While ICoref-\textit{inc} does not rely on speaker embeddings, our own models (both Part-Inc and Sent-Inc) do.
Given the important role of speaker identity in a dialogue setting, it is useful to know the effect of removing these embeddings in our models.

We compare the Sent-Inc model with and without speaker embeddings in Table \ref{tab:ablation_results} for OntoNotes, and Table \ref{tab:codi_ablation} for CODI-CRAC.
We find that speaker embeddings play little to no role in coreference performance.
In OntoNotes, removing speaker embeddings improves CoNLL F1 by 0.1, and in CODI-CRAC, it decreases performance by 0.2 F1.
In both cases, the results are unlikely to be statistically significant.
The finding indicates that Sent-Inc's advantage over ICoref-\textit{inc} is not simply due to feature selection but a true modelling advantage.
It also suggests that further performance improvements are possible if speaker identity can be better represented, since Sent-Inc effectively ignores the speaker embeddings.
One possibility, from \citet{wu-etal-2020-corefqa}, is to preprocess the text with speaker tags directly included in the input, rather than including it separately.
This way, the document encoder directly learns how to handle speakers, instead of relying on a separate embedding in downstream classifiers.

\begin{table*}[ht]
\small
\centering
\begin{tabular}{l|c c c c c c c c c|c}
\toprule		
	&		\multicolumn{3}{c}{\muc}					&	\multicolumn{3}{c}{\bcub}					&	\multicolumn{3}{c}{\ceaf}					&	Avg.	\\
Model	&	Rec.	&	Prec.	&	F1	&	Rec.	&	Prec.	&	F1	&	Rec.	&	Prec.	&	F1	&	F1	\\
\midrule
SpanBERT+\textit{e2e-coref} & 82.5	&	84.3	&	83.4	&	75.8	&	76.8	&	76.3	&	71.7	&	74.7	&	73.1	&	77.6	\\
XLNet+\textit{e2e-coref} & 70.3	&	80.8	&	75.2	&	63.7	&	73.0	&	68.0	&	67.5	&	70.4	&	68.9	&	70.7	\\
\bottomrule								
\end{tabular}
\caption{Results from non-incremental methods on the OntoNotes dev set with different pretrained language models. \textit{e2e-coref} refers to the non-incremental formulation from \citet{lee-etal-2018-higher}, which is adapted in \citet{joshi-etal-2020-spanbert}.
}
\label{tab:xlnet_ablation}
\end{table*}

\section{XLNet in Non-Incremental Baselines}

Choosing XLNet as the document encoder is motivated by the fact that XLNet can efficiently cache and reuse input, making it suitable for incremental processing.
However, XLNet can also be used non-incrementally in the same way as SpanBERT.
In particular, we can train \citet{joshi-etal-2020-spanbert}'s coreference system using XLNet instead of SpanBERT.
This experiment allows us to compare how the choice of pre-trained language model affects performance.

Table \ref{tab:xlnet_ablation} shows results of training \citet{joshi-etal-2020-spanbert} with an XLNet encoder instead of SpanBERT on the OntoNotes dev set. 
XLNet significantly underperforms compared to SpanBERT, scoring almost 7 CoNLL F1 points lower.
Surprisingly, XLNet is an effective document encoder for our Part-Incremental formulation (achieving 76.3 F1 on the OntoNotes test set), but ineffective when used in \citet{joshi-etal-2020-spanbert}'s non-incremental setup.
We do not attempt swapping the fine-tuned XLNet into \citet{toshniwal-etal-2020-learning} or \citet{xia-etal-2020-incremental} as it seems unlikely to yield useful results.

\section{Hyperparameters and Other Model Details}

The main hyperparameters are listed in Table \ref{tab:hyperparam}. 

\begin{table}[ht]
    \centering
    \begin{tabular}{l|c}
\toprule			
Hyperparameter	&	Value	\\ \midrule
Encoder Learning Rate	&	2e-5	\\
Task Learning Rate	&	1e-4	\\
Adam Eps	&	1e-6	\\
Adam Weight Decay	&	1e-2	\\
Gradient Clipping Norm	&	1	\\
Dropout Rate	&	0.3	\\
$\stacklstm$ Hidden Size	&	200	\\
Action History Hidden Size	&	30	\\
$f_M$ Hidden Size	&	1000	\\
$f_C$ Hidden Size    &   3000 \\
New Cell Threshold ($\alpha$)	&	0.0	\\
$\phi_C$ Max Entity Count	&	10	\\
$\phi_C$ Max Mention Distance	&	10	\\
$\phi_M$ Max Span Width	&	30	\\
Max Speaker Number	&	20	\\
\bottomrule	
    \end{tabular}
    \caption{Hyperparameters used during training.}
    \label{tab:hyperparam}
\end{table}

The bottom four rows refer to the maximum number of learned embeddings we use for each feature. Additionally:
\begin{itemize}
    \item The top performing Part-Inc model uses 20 sentences as active input, with the remainder as memory (up to 512 tokens total).
    \item During training, the Sent-Inc model accumulates gradients after every 32 sentences to ensure that the memory used does not exceed capacity.
\end{itemize}

Our implementation is based off of \citet{xu-choi-2020-revealing}'s codebase.
We find their model hyperparameters are already extremely well-tuned, and so we do not explore further hyperparameter tuning for these cases.
Regarding new hyperparameters introduced in this work, we follow previous work in choosing sensible values.
For example, the $\stacklstm$ and Action History LSTM hidden sizes follow \citet{dyer-etal-2015-transition}'s recommendations.

We train all models using NVIDIA Tesla V100 16 GB cards on an HPC cluster. 
Training convergence takes approximately 24 hours.
Both Sent-Inc and Part-Inc models contain around 140 million parameters.

\section{Dataset Details}

For all datasets, we follow standard preprocessing steps such as tokenization, mapping subword units to token IDs, and adding segment boundary tokens (such as [CLS] and [SEP]).
Since our algorithms rely on teacher forcing, we compute gold actions for both mention detection and mention clustering steps.

The English portion of OntoNotes 5.0 contains 3493 documents, divided into 2802/343/348 splits for training, validation and test sections, respectively.
The corpus contains examples for English, Chinese and Arabic; however, we only use the English portion in this paper.
Details to download the dataset can be found at: \href{https://catalog.ldc.upenn.edu/LDC2013T19}{https://catalog.ldc.upenn.edu/LDC2013T19}.
The official scorer is available at \href{https://github.com/conll/reference-coreference-scorers}{https://github.com/conll/reference-coreference-scorers}.
We do not remove data from OntoNotes.

The CODI-CRAC 2021 corpus contains 134 English documents, split into 60 documents for validation and 74 documents for testing.
Since we use the corpus solely for evaluation, we only use the test set.
The corpus is available at: \href{https://codalab.lisn.upsaclay.fr/competitions/614}{https://codalab.lisn.upsaclay.fr/competitions/614}, although a license from the Linguistic Data Corsortium (LDC) is needed for the Switchboard portion.
The official scorer can be found at: \href{https://github.com/juntaoy/universal-anaphora-scorer}{https://github.com/juntaoy/universal-anaphora-scorer}.
As detailed in Section \ref{sec:datasets}, we remove singletons due to lack of training data.

\section{Evaluation Metrics}

As mentioned in Section \ref{sec:experiments}, we evaluate using the \muc, \bcub\ and \ceaf metrics, together with their average -- commonly referred to as the CoNLL score after its usage in the CoNLL 2012 Shared Task \cite{pradhan-etal-2012-conll}.

None of the metrics have a trivial definition.
Each metric measures different aspects of the predicted coreference chains.
We do not provide a full definition of each metric here, but note that \muc\ corresponds to a \textit{link}-based measure, \bcub\ is a \textit{mention}-based measure, and \ceaf\ is an \textit{entity}-based measure.
We refer the reader to the CoNLL 2012 Shared Task paper \cite{pradhan-etal-2012-conll} for an overview of all three metrics and to the original papers for a full description \cite{vilain-etal-1995-model, bagga1998algorithms, luo-2005-coreference}.

\end{document}